\documentclass[letterpaper]{article}

\usepackage[margin=1.5in]{geometry}

\usepackage[utf8]{inputenc}
\usepackage[cmex10]{amsmath}
\usepackage{amssymb}
\usepackage[hang,small]{caption}
\usepackage[dvips]{graphicx}
\usepackage{color}
\usepackage{subfig}
\usepackage{theorem}
\usepackage{bbold}
\usepackage{cite}

\usepackage{hyperref}

\newcommand{\RR}{\ensuremath{\mathbb R}}
\usepackage{algorithmic}
\usepackage[ruled,vlined]{algorithm2e}

\DeclareMathOperator{\sgn}{sgn}
\newcommand{\constr}{z}

\newcommand{\HBx}{\widehat{\mathbf{x}}}
\newcommand{\HBu}{\widehat{\mathbf{u}}}
\newcommand{\HBz}{\widehat{\mathbf{z}}}

\newcommand{\Hbx}[1]{\boldsymbol{\widehat{x}}_{#1}}
\newcommand{\Hbu}[1]{\boldsymbol{\widehat{u}}_{#1}}

\newcommand{\Bx}{\mathbf{x}}
\newcommand{\Bu}{\mathbf{u}}
\newcommand{\By}{\mathbf{y}}
\newcommand{\Bz}{\mathbf{\constr}}
 
\newcommand{\bx}[1]{\boldsymbol{x}_{#1}}
\newcommand{\bu}[1]{\boldsymbol{u}_{#1}}
\newcommand{\by}[1]{\boldsymbol{y}_{#1}}

\newtheorem{theorem}{Theorem}[section]
\newtheorem{proposition}[theorem]{Proposition}
\theoremstyle{plain}{\theorembodyfont{\rmfamily}%
}
\theoremstyle{plain}{\theorembodyfont{\rmfamily}%

\newtheorem{remark}[theorem]{Remark}}

\graphicspath{
{./figures/}
}

\begin{document}

\title{On-the-fly Approximation of\\ Multivariate Total Variation Minimization}
\date{}

\author{Jordan~Frecon\thanks{J. Frecon  (Corresponding author), N. Pustelnik and P. Abry are with the Physics Department of the ENS Lyon in France (email: firstname.lastname@ens-lyon.fr).}, Nelly~Pustelnik$^*$, Patrice~Abry$^*$, and~Laurent~Condat\thanks{L. Condat is with the GIPSA-lab, University of Grenoble, France (email: laurent.condat@gipsa-lab.grenoble-inp.fr).}
\vspace{-0.01cm}}

\maketitle

\begin{abstract}
In the context of change-point detection, addressed by Total Variation minimization strategies, an efficient on-the-fly algorithm has been designed leading to exact solutions for univariate data. 
In this contribution, an extension of such an on-the-fly strategy to multivariate data is investigated. 
The proposed algorithm relies on the local validation of the Karush-Kuhn-Tucker conditions on the dual problem.
Showing that the non-local nature of the multivariate setting precludes to obtain an exact on-the-fly solution, 
we devise an on-the-fly algorithm delivering an approximate solution, whose quality is controlled by a practitioner-tunable parameter, acting as a trade-off between quality and computational cost. Performance assessment shows that high quality solutions are obtained on-the-fly while benefiting of computational costs several orders of magnitude lower than standard iterative procedures.
The proposed algorithm thus provides practitioners with an efficient multivariate change-point detection on-the-fly procedure. 
\end{abstract}

\clearpage

\section{Introduction}
Total Variation (TV) has been involved in a variety of signal processing problems, such as nonparametric function estimation \cite{Mammen_E_1997_j-anns_locally_ars, Davies_P_2001_j-annals-statistics_le_rsm} or signal denoising \cite{Harchaoui_Z_2008_p-nips_catching_cplasso, Vert_JP_2010_p_nips_fast_dmcpsmsug, Condat_L_2013_j-ieee-spl_direct_a1D}. The first contributions on this subject were formulated within the framework of taut string theory \cite{Mammen_E_1997_j-anns_locally_ars,Davies_P_2001_j-annals-statistics_le_rsm} while the term TV had first been introduced in image restoration~\cite{Rudin_L_1992_j-phys-d_tv_atvmaopiip, Chambolle_A_2010_book_itvia}. The equivalence between both \mbox{formalisms has been clarified in \cite{Grasmair_M_2007_j-math-imaging-vis_equivalence_tsabvr}.} 

Formally, the univariate TV framework aims at finding a piece-wise constant estimate $\widehat{\bx{}} \in \RR^N$ of a noisy univariate signal $\boldsymbol{y}\in \RR^N$ by solving the following non-smooth convex optimization problem,
\begin{equation}
\label{eq:univariatetv1D}
\widehat{\boldsymbol{x}} = \underset{\boldsymbol{x} = (x_k)_{1\leq k \leq N}}{\arg\min} \frac{1}{2} \Vert \boldsymbol{x} - \boldsymbol{y} \Vert^2 + \lambda \sum_{k=1}^{N-1} |x_{k+1}-x_k|, 
\end{equation}
where $\lambda>0$ denotes a regularization parameter balancing data fidelity versus the piece-wise nature of the solution.\\

\noindent {\bf Related works: recent developments and issues.}~ 
It is well known and documented that the unique solution of the optimization problem \eqref{eq:univariatetv1D} can be reached by iterative fixed-point algorithms. On the one hand, solving this problem in the primal space requires to deal with the non-differentiability of the $\ell_1$-norm that is either handled by adding a small additional smoothing parameter \cite{Vogel_C_2002_book_com_mip} or by considering proximal algorithms \cite{ Condat_L_2013_j-ieee-spl_direct_a1D, Steidl_G_2006_j-ijcv_splines_hot, Combettes_PL_2008_j-ip_proximal_apdmfscvip, Combettes_P_2010_inbook_proximal_smsp, Afonso_M_2009_j-tip_augmented_lacofiip, Chambolle_A_2010_first_opdacpai,Pesquet_J_2012_j-pjpjoo_par_ipo, Rinaldo_A_2009_properties_rflasso, Bach_F_2012_j-ftml_optimization_sip, Bauschke_H_2011_book_con_amo, Alais_C_2013_book_gfl}. On the other hand, one can make use of the  Fenchel-Rockafellar dual formulation~ \cite{Chambolle_A_2004_jmiv_TV_aaftvmaa,Steidl_G_2006_j-ijcv_splines_hot} or Lagrangian duality~\cite{Kim_S_j-ieee-jstsp_interior_pml, Wahlberg_B_2012_ifac_admmactvrep} that can be solved with quadratic programming techniques \cite{Steidl_G_2006_j-ijcv_splines_hot, Barbero_A_2011_p-icml_fntmtvr}. Both primal and dual solutions suffer from high computational loads, stemming from their iterative nature. To address the computational load issue, alternative procedures were investigated, such as the \emph{taut string algorithm} of common use in the statistics literature \cite{Mammen_E_1997_j-anns_locally_ars}. Very recently, elaborating on the dual formulation and thoroughly analysing the related Karush-Kuhn-Tucker (KKT) conditions, a fast algorithm has been proposed by one of the authors in \cite{Condat_L_2013_j-ieee-spl_direct_a1D} to solve the univariate optimization problem~\eqref{eq:univariatetv1D}. Compared to the taut string strategy, it permits to avoid running sum potentially leading to overflow values and thus numerical errors. Another specificity concerns 
its \emph{on-the-fly} behavior that does not require the observation of the whole time sequence before a solution can be obtained. On-the-fly algorithms might be of critical interest for real-time monitoring such as in medical applications~\cite{Godinez_M_2003_p-embs_onl_fhr,Cannon_J_2010_j-bspc_alg_odt}.

Along another line, extension of the univariate optimization problem~\eqref{eq:univariatetv1D} to multivariate purposes has been recently investigated in~\cite{Vert_JP_2010_p_nips_fast_dmcpsmsug, Bleakley_K_2011_techrep_group_flm, Hinterberger_W_2003_j-math-imaging-vis_tmbvr}. The multivariate extension arises very naturally in numerous contexts, such as biomedical applications, for which the purpose is to extract simultaneous change points from multivariate data, e.g., EEG data~\cite{Gramfort_A_2012_j-pmb_mix_nem}. It also encompasses denoising of complex-valued data, which can naturally be interpreted as bivariate data. Multivariate optimization is known as the group fused Lasso in the statistics literature \cite{Yuan_M_2006_j-r-stat-s-b_model_ser, Liu_J_2013_ACM_eacflp}. From a Bayesian point of view, elegant solutions have been proposed in \cite{Dobigeon_N_2007_j-ieee-tsp_joi_sma, Picard_2005_BMCBioinformatics_statistical_aaCGHda} and efficient iterative strategies have recently been proposed in~\cite{Bleakley_K_2011_techrep_group_flm, Hoefling_H_2009_arxiv_path_aflassosa}.
\vskip\baselineskip
\clearpage
\noindent {\bf Mutivariate on-the-fly TV.}~ In this context, the present contribution elaborates on \cite{Condat_L_2013_j-ieee-spl_direct_a1D} to propose an on-the-fly algorithm solving the multivariate extension of \eqref{eq:univariatetv1D}.
In Section~\ref{sec:localVsGlobal}, the group fused Lasso problem is first detailed. It is then illustrated that the multivariate procedure has a \emph{non-local} behavior as opposed to the \emph{local} nature of the univariate problem~\eqref{eq:univariatetv1D}. Consequently, any on-the-fly algorithm solving the multivariate minimization problem will only lead to an approximate solution. 
The KKT conditions resulting from the dual formulation of the multivariate problem are specified in Section~\ref{sec:bivTV}. From such conditions, a fast and on-the-fly, yet approximate, algorithm is derived in Section~\ref{sec:algoSol}. The performance in terms of achieved solution and computational gain are presented in Section~\ref{sec:res}. A video demonstrating the on-the-fly behavior of the algorithm as well as a \textsc{Matlab} toolbox are available at \url{http://perso.ens-lyon.fr/jordan.frecon}.\\

\noindent\textbf{Notations.}~ Let $\Bu = (u_{m,k})_{1\leq m \leq M, 1\leq k \leq N}\in\mathbb{R}^{M\times N}$ denote a multivariate signal, where for every $m\in\{1,\ldots,M\}$, $\bu{m} = (u_{m,k})_{1\leq k \leq N}\in\mathbb{R}^N$ stands for the $m$-th component while the $k$-th values will be shortened as $\Bu_{k}= (u_{m,k})_{1\leq m \leq M}\in\mathbb{R}^M$. For every $k\in\{1,\ldots,N\}$, use will also be made of the following functions: $\mathrm{abs}(\Bu_{k}) = (\vert u_{m,k}\vert)_{1\leq m \leq M}\in\mathbb{R}^M$ and $\sgn(\Bu_{k})=\big( \sgn(u_{m,k})\big)_{1\leq m \leq M}\in\mathbb{R}^M$.

\section{\label{sec:localVsGlobal}Local vs non-local Nature}

We denote $\By$ the multivariate signal of interest. 
A multivariate extension of \eqref{eq:univariatetv1D} reads:
\begin{equation}
\label{eq:pbPrimal1}
\small
\HBx =  \underset{{\Bx} = (\bx{1},\ldots,\bx{M})}{\arg\min} \;\frac{1}{2}\sum_{m=1}^M\|\bx{m}-\by{m}\|^2 + \lambda\sum_{k=1}^{N-1}\sqrt{\sum_{m=1}^M  |(L\bx{m})_k|^2 },
\end{equation}
where $\lambda> 0$ denotes the regularization parameter and  $L\in\RR^{(N-1)\times N}$ denotes the first order difference operator, that is, for $m\in\{1,\ldots,M\}$ and $k\in\{1,\ldots, N-1\}$, 
\begin{equation}(L\bx{m})_k = x_{m,k+1}-x_{m,k}.\end{equation}

\begin{figure}[t]
\centering\includegraphics[scale=.52, clip=true, trim=1cm 1cm 0cm .7cm]{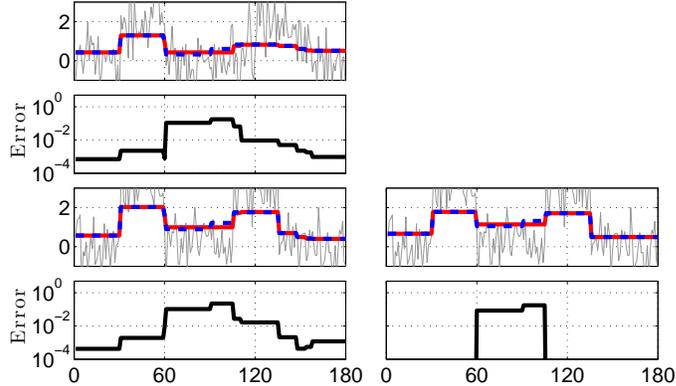}
\caption{\label{fig:localVsGlobal}\textbf{Non-local vs. local nature.} Left: bivariate TV (upper plots: first component, lower plots: second component). Right: univariate TV. Observations $\By$ (gray), solution $\HBx$ (red), concatenation of solutions $\HBx_-$ and $\HBx_+$ (dashed blue).}
\end{figure}

Despite formal similarity, there is however a fundamental difference in nature between the univariate ($M=1$) and multivariate ($M>1$) cases: 
The former is intrinsically \textit{local}~\cite{ Condat_L_2013_j-ieee-spl_direct_a1D, Louchet_C_2011_j-siam-is_tvlf} while the latter is \textit{non-local}\footnote{In this article, we denote a problem as local if the solution at a given location does not depend on the signal located earlier (later) than the previous (next) change-point.}. 
To make explicit such a notion, we have designed the following experiment, whose results are illustrated in Fig.~\ref{fig:localVsGlobal}. The results associated to the univariate (resp. bivariate) case are presented on the right plots (resp. left plots).
A univariate signal $\by{}\in \mathbb{R}^N$ with $N=180$, consisting of the additive sum of a piece-wise constant signal and white Gaussian noise (in gray, in Fig.~\ref{fig:localVsGlobal}, right top plot), is considered first. 
The solution of the minimization problem~\eqref{eq:univariatetv1D} is displayed in solid red lines in Fig.\ref{fig:localVsGlobal}. 
Also, we search for the solution of the minimization problem~\eqref{eq:univariatetv1D} applied to two partitions of $\by{}$, 
obtained by splitting it in half, i.e., $\by{-}=(y_{k})_{1\leq k \leq N/2}$ and $\by{+}=(y_{k})_{N/2+1\leq k \leq N}$.
The solutions $\bx{-}$ and $\bx{+}$ of \eqref{eq:univariatetv1D} respectively associated to $\by{-}$ and $\by{+}$ are concatenated and displayed with dashed blue lines in Fig.~\ref{fig:localVsGlobal}. 
There is strictly no difference between $\bx{}$ and the concatenation of $\bx{-}$ and $\bx{+}$, as reported in Fig.\ref{fig:localVsGlobal} (bottom right plot), except for the segment that contains the concatenation point. 
The difference around the concatenation point is expected as $\bx{}$ makes use of an information (the continuity between $\by{-}$ and $\by{+}$) that is not available to compute $\bx{-}$ and $\bx{+}$. 
The fact that there is no difference elsewhere shows the local nature of the univariate solution to Problem~\eqref{eq:univariatetv1D}.

This experiment is now repeated for $M= 2$ (as the simplest representative of $M >1$). 
A bivariate signal ${\By}=(\by{1},\by{2})\in\mathbb{R}^{2\times N}$ with $N=180$, consisting of the additive sum of piece-wise constant signals and white Gaussian noises (in gray, in Fig.~\ref{fig:localVsGlobal}, left plots, 1st and 3rd lines), is considered. 
Two partitions, obtained by splitting in half, ${\By}_-=(y_{1,k},y_{2,k})_{1\leq k \leq N/2}$ and ${\By}_+=(y_{1,k},y_{2,k})_{N/2+1\leq k \leq N}$ are also considered. 
The corresponding solutions of \eqref{eq:pbPrimal1}, applied to ${\By}, {\By}_-, {\By}_+$, labeled $\HBx$, $\HBx_-$ and $\HBx_+$ are obtained by means of the primal-dual algorithm proposed in~\cite{Chambolle_A_2010_first_opdacpai} with  $\lambda=20$. 
Solutions $\HBx_-$ and $\HBx_+$ of \eqref{eq:pbPrimal1} respectively associated to ${\By}_-$ and ${\By}_+$ are concatenated and displayed with dashed blue lines in Fig.~\ref{fig:localVsGlobal}, while $\HBx$ is shown in red. 
Contrary to the case $M=1$, differences between $\HBx$ and concatenated $\HBx_-$ and $\HBx_+$, shown in black in bottom plots, differ unambiguously from $0$ over the entire support of ${\By}$, clearly showing the \emph{non-local} nature of $\HBx$ when $M > 1$. 

In the univariate case (Eq.~\eqref{eq:univariatetv1D}), the \emph{local} nature of the solution permits to design an efficient taut string algorithm, that consists in finding the string of minimal length (i.e., taut string) that holds in the tube of radius $\lambda$ around the antiderivative of $\mathbf{y}$.
The solution $\widehat{\boldsymbol{x}}$ of \eqref{eq:univariatetv1D} is then obtained by computing the derivative of the taut string. 
An efficient strategy has been proposed in~\cite{Davies_P_2001_j-annals-statistics_le_rsm} in order to straightforwardly compute $\widehat{\boldsymbol{x}}$ by determining the points of contact between the taut string and the tube. 
Even though this approach can be generalized to multivariate signals, the detection of points of contact additionally requires the angle of contact between the taut string and the tube. However, this information is \emph{non-local} and thus the on-the-fly minimization problem results in a challenging contact problem which can not be solved locally. 
This interpretation will be further discussed in Section~\ref{sec:bivTV}.

The \emph{non-local} nature of the multivariate ($M>1$) Problem~\eqref{eq:pbPrimal1} implies that one cannot expect to find an exact multivariate on-the-fly algorithm. 
Therefore, in the present work, we will derive an \textit{approximate} on-the-fly algorithm that provides us a good-quality approximation of the exact solution to Problem \eqref{eq:pbPrimal1}. 
A control parameter $| \mathcal{Q} |$, defined in Section 3, will control the trade-off between the quality of the approximation and the computational cost. 
\section{Multivariate Total Variation Minimization}
\label{sec:bivTV}
\subsection{Dual formulation}
\begin{figure}[t]
\centering
\includegraphics[scale=.35]{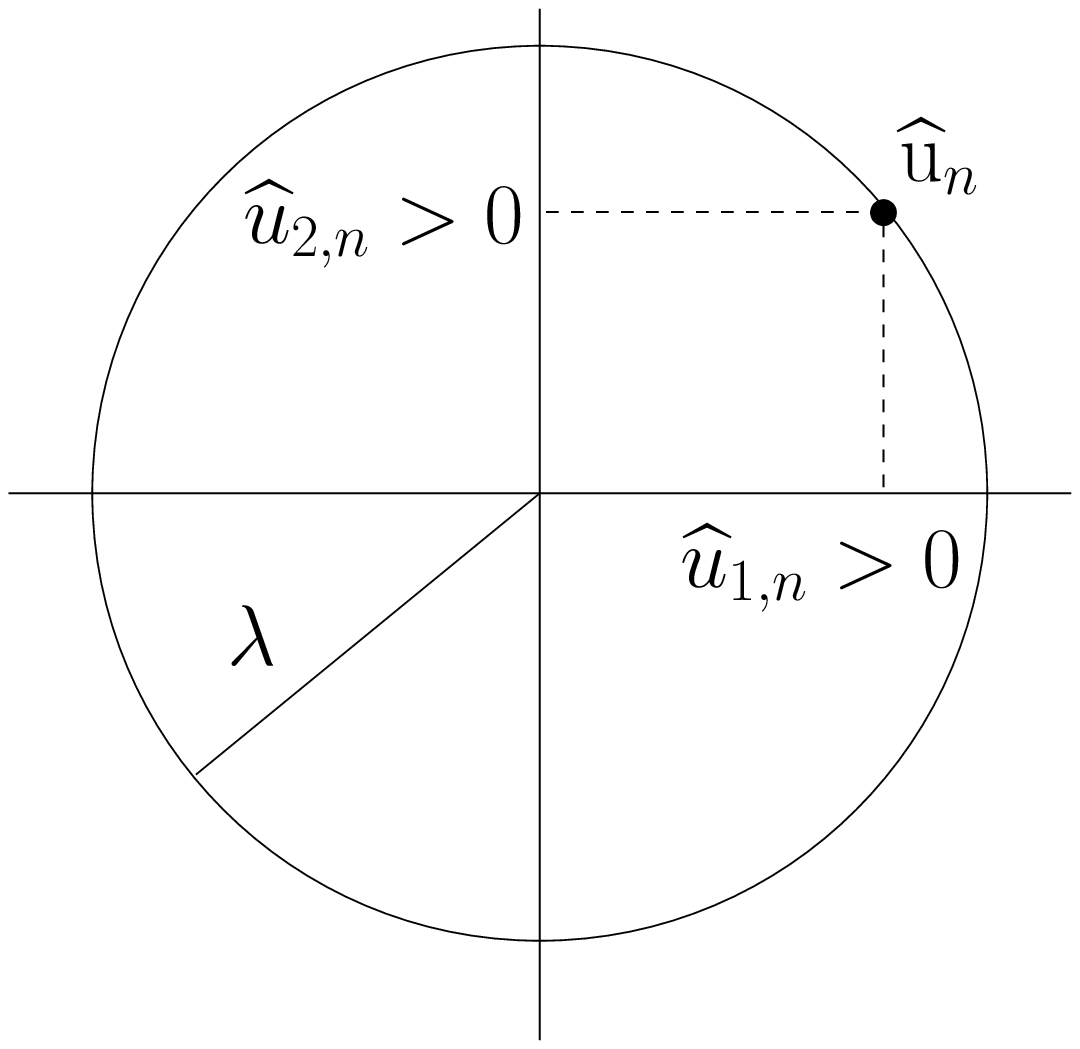}\qquad
\includegraphics[scale=.35]{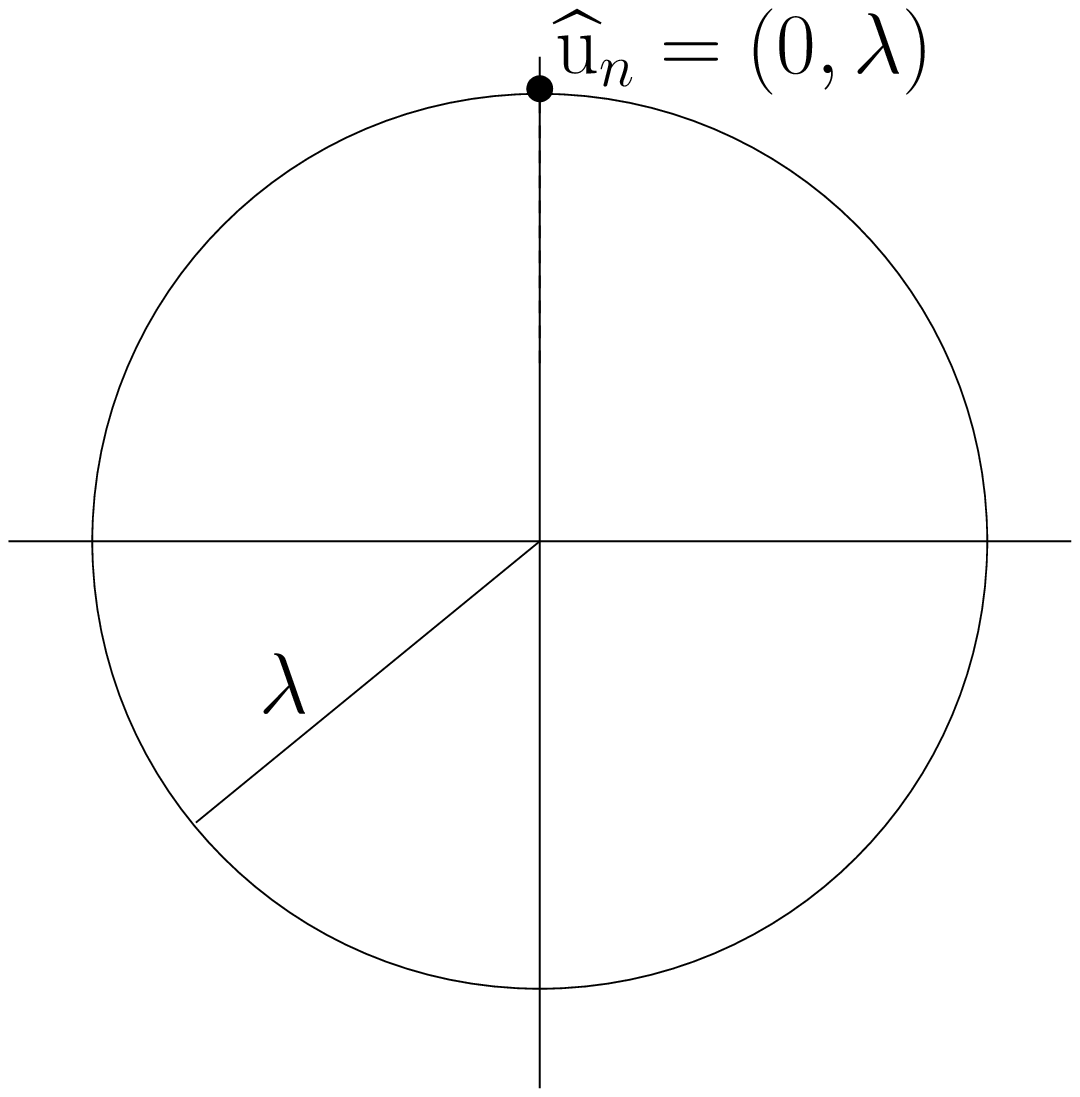}
\caption{\label{fig:illustrDual}\textbf{Comparing joint vs disjoint changes in the dual space.} Left: location $k$ is suitable for a joint negative amplitude change on both components. Right: configuration suitable for introducing a negative amplitude change at $k$ on the second component only.}
\end{figure}

Fenchel-Rockafellar dual formulation\footnote{Note that, the usual dual formulation and the resulting stationarity conditions would involve $\Bu$ rather than $-\Bu$. The choice made in this article enables us to be consistent with the results obtained in \cite{Condat_L_2013_j-ieee-spl_direct_a1D} for the univariate case.} of~\eqref{eq:pbPrimal1} reads :
\begin{equation}
\!\!\!\underset{ \Bu \in \RR^{M\times (N-1)}}{\text{minimize}}\; \frac{1}{2}\sum_{m=1}^M\|\by{m}+L^*\bu{m}\|^2 \quad \text{subject to}\quad
(\forall k=\{1,\ldots, N-1\}) \quad \| \Bu_k \|\leq\lambda,
\label{eq:pbDualConstr1}
\end{equation}
where, for every $ m\in\{1,\ldots,M\}$ and $k=\{2,\ldots, N-2\}$,
\begin{equation}
(L^*\widehat{\boldsymbol{u}}_m)_k = \widehat{u}_{m,k-1} -\widehat{u}_{m,k}
\end{equation}
and 
\begin{equation}
\begin{cases}
(L^*\widehat{\boldsymbol{u}}_m)_1 &= - \widehat{u}_{m,1}, \\
(L^*\widehat{\boldsymbol{u}}_m)_N &=  \widehat{u}_{m,N-1}. \\
\end{cases}
\end{equation}
The optimal solutions $\HBu \in\mathbb{R}^{M\times(N-1)}$ and  $\HBx \in \RR^{M\times N}$ of the dual problem and of the primal problem respectively are related by
\begin{equation}
\label{eq:optPrimalDual}
\!\!\!\!\begin{cases}
(\forall m\in\{1,\ldots,M\})&\Hbx{m}=\by{m}+{L}^*\Hbu{m},\\
(\forall k\in\{1,\ldots,N-1\}) &\HBu_{k} \in -\lambda \partial \Vert \cdot \Vert (\HBx_{k+1} - \HBx_{k}).
\end{cases}
\end{equation}
From \eqref{eq:optPrimalDual}, we directly obtain the following necessary and sufficient conditions.
\vspace{-0.1cm}
\begin{proposition}
\label{th:FOC}
The solutions of the primal problem~\eqref{eq:pbPrimal1} and the dual problem~\eqref{eq:pbDualConstr1} satisfy, for every  $m\in\{1,\ldots,M\}$,
\begin{equation}
\Hbx{m}=\by{m}+{L}^*\Hbu{m},
\end{equation}
and, for every $k=\{1,\ldots,N-1\}$,
\begin{equation}
\label{eq:KKT}
\begin{cases} 
\mbox{if} \quad \HBx_k = \HBx_{k+1} \quad \mbox{then} \quad  \Vert\HBu_k\Vert \leq \lambda,\\
\mbox{if} \quad \HBx_k \neq \HBx_{k+1} \quad \mbox{then}\quad \HBu_k = -\lambda \frac{\HBx_{k+1}-\HBx_{k}}{\Vert \HBx_{k+1}-\HBx_{k} \Vert}.
\end{cases}
\end{equation}
\end{proposition}
The first condition corresponds to the configuration where every component keeps the same value from location $k$ to \mbox{$k+1$}. This configuration is illustrated in the bivariate case ($M=2$) in Fig.~\ref{fig:illustrDual} (left plot). The second condition models situations where some components of $\HBx$ admit change points between locations $k$ and $k+1$. An interesting configuration is that of non-simultaneous change points as illustrated in Fig.~\ref{fig:illustrDual} (right plot). In the presence of noise, this second situation is rarely encountered. Thus, in the sequel, we will only consider simultaneous change points.

\vspace{-0.1cm}
\begin{remark}
Proposition~\ref{th:FOC} for $M=1$ leads to the usual KKT conditions associated to the minimization problem~\eqref{eq:univariatetv1D}:
\begin{equation}
\label{eq:1DOC}
\begin{cases} 
\mbox{if} \quad \widehat{x}_{k} > \widehat{x}_{k+1}\quad \mbox{then} \quad  \widehat{u}_k = + \lambda,\\
\mbox{if} \quad \widehat{x}_{k} < \widehat{x}_{k+1} \quad \mbox{then} \quad \widehat{u}_k = - \lambda,\\ 
\mbox{if} \quad \widehat{x}_{k} = \widehat{x}_{k+1}  \quad \mbox{then} \quad \widehat{u}_k \in [-\lambda, +\lambda].
\end{cases}
\end{equation}
\end{remark}
The on-the-fly univariate TV algorithm proposed in~\cite{Condat_L_2013_j-ieee-spl_direct_a1D} is derived from Conditions~\eqref{eq:1DOC}. 
\subsection{Rewriting the KKT conditions}
Contrary to Conditions~\eqref{eq:1DOC}, the multivariate conditions derived in Proposition~\ref{th:FOC} are not directly usable in practice to devise an on-the-fly algorithm because $\HBx_{k+1} - \HBx_{k}$ is a priori unknown at instant $k$. Therefore, we propose to rewrite the second condition in \eqref{eq:KKT} by means of auxiliary variables $(\HBz_{k})_{1\leq k \leq N-1}$ such that
\begin{equation}
\begin{cases} 
\mbox{if} \quad \HBx_k = \HBx_{k+1} \quad \mbox{then} \quad  \Vert\HBu_k\Vert \leq \lambda,\\
\mbox{if} \quad \HBx_k \neq \HBx_{k+1} \quad \mbox{then}\quad \HBu_k = - \mathrm{sign}(\HBx_{k+1}-\HBx_{k})\circ\HBz_{k},
\end{cases}
\end{equation}
with $\HBz_{k} = \lambda\frac{\mathrm{abs}(\HBx_{k+1}-\HBx_{k})}{\Vert \HBx_{k+1}-\HBx_{k} \Vert}$ and where $\circ$ denotes the Hadamard product. 
Then, Proposition~\ref{th:FOC}, can be reformulated component-wise as follows.
\begin{proposition}
\label{th:FOC2}
The solutions of the primal problem~\eqref{eq:pbPrimal1} and of the dual problem~\eqref{eq:pbDualConstr1} satisfy the following necessary and sufficient conditions. There exist nonnegative auxiliary variables $(\HBz_{k})_{1\leq k \leq N-1}$ such that, for every $m=\{1,\ldots,M\}$ and  $k=\{1,\ldots,N-1\}$, 
\begin{equation}
\label{eq:2DOC}
\begin{cases} 
\mbox{if} \quad \widehat{x}_{m,k} > \widehat{x}_{m,k+1}\quad \mbox{then} \quad  \widehat{u}_{m,k} = + \widehat{\constr}_{m,k},\\
\mbox{if} \quad \widehat{x}_{m,k} < \widehat{x}_{m,k+1} \quad \mbox{then} \quad \widehat{u}_{m,k} = - \widehat{\constr}_{m,k},\\ 
\mbox{if} \quad \widehat{x}_{m,k} = \widehat{x}_{m,k+1}  \quad \mbox{then} \quad \widehat{u}_{m,k} \in [-\widehat{\constr}_{m,k}, +\widehat{\constr}_{m,k}],
\end{cases}
\end{equation}
with $\|\HBz_{k}\| = \lambda$ and  $\Hbx{m}=\by{m}+{L}^*\Hbu{m}$.
\end{proposition}

Comparing Eqs.~\eqref{eq:1DOC} and \eqref{eq:2DOC} highlights the similarity between the necessary conditions of the univariate and multivariate minimization problems: Conditions involving $\lambda$ in the univariate case involve the auxiliary vector $\HBz{}$ in the multivariate one. 
The fact that $\HBz{}$ differs for each pair $(m,k)$ can be interpreted in taut string procedures as the fact that the point of contact with the taut string may vary on the tube of radius $\lambda$. This significantly increases the difficulty of deriving an on-the-fly algorithm.
\subsection{Approximate solution}
\label{ssec:approx}
If we first assume that $\HBz{}$ is known and such that, for every $k\in\{1,\ldots,N-1\}$, $\| \HBz_k \|=\lambda$, the primal problem associated to Conditions~\eqref{eq:2DOC} reads
\begin{equation}
\!\!\!\underset{\Bx{}}{\min}\; \sum_{m=1}^M \left(\frac{1}{2}\|\by{m}-\bx{m}\|^2 + \sum_{k=1}^{N-1} \widehat{z}_{m,k} | (L \bx{m})_k | \right)
\end{equation}
and can be interpreted as $M$ univariate minimization problems having  time-varying regularization parameters $(\widehat{\boldsymbol{\constr}}_{m})_{1\leq m \leq M}$. 

The proposed approximation consists in restricting the estimation of $\HBz$ to a predefined set $\mathcal{Q}=\{\zeta^{(1)},\ldots,\zeta^{(|\mathcal{Q}|)}\}$ chosen such as for every $q\in\{1,\ldots,|\mathcal{Q}|\}$, $\zeta^{q}=(\zeta_m^{(q)})_{1\leq m \leq M} \in \RR^M$ satisfies $\| \zeta^{(q)} \| = \lambda$.

The most naive strategy would consist in solving $M$ univariate minimization problems for every $|\mathcal{Q}|$ candidate values of $\HBz$, i.e., find for every  $m=\{1,\ldots, M\}$ and $q=\{1,\ldots, |\mathcal{Q}|\}$,\\
\begin{equation}
\widehat{\bx{}}_m^{(q)} = \mathrm{arg} \min_{\bx{m}} \frac{1}{2}\|\by{m}-\bx{m}\|^2 + \zeta_m^{(q)} \| L\bx{m} \|_1
\end{equation}
and to devise a method to pick the solution amongst the $|\mathcal{Q}|$ candidates. For instance, the one that maximizes some quality criterion $f$, i.e.,
\begin{equation}
\HBx=\HBx^{(q^*)}\quad \mbox{with} \quad q^* = \mathrm{arg} \max_{1\leq q \leq |\mathcal{Q}|} f( \HBx^{(q)}).
\end{equation}
Although it benefits from parallel on-the-fly implementations, this situation would correspond to a constant estimate $ \tilde{\Bz} = \zeta^{(q^*)}\mathbb{1}_N$.
Therefore, changes in the mean would be processed independently on all components and group-sparsity would not be enforced.

In order to benefit from an on-the-fly implementation and to enforce group-sparsity, we propose an algorithmic solution based on a piece-wise constant estimator of $\HBz$ detailed in the next section.
\section{Algorithmic solution}
\label{sec:algoSol}

In the following, we first extend the on-the-fly algorithm proposed in \cite{Condat_L_2013_j-ieee-spl_direct_a1D} to the multivariate case, with $\HBz$ assumed to be known a priori. This strong assumption, unrealistic in pratice, permits to describe clearly the behaviour of the multivariate on-the-fly algorithm. Then, we will focus on the question of the automated and on-the-fly estimation of $\HBz{}$  taking its values in $\mathcal{Q}$, which consequently introduce a parameter $|\mathcal{Q}|$ controlling the quality of the approximation. The main steps of the on-the-fly algorithm are summarized in Algorithm~\ref{algo:scheme}. It is based on the range control of both unknown primal and dual solutions $\widehat{\Bx}$ and $\widehat{\Bu}$ by lower and upper bounds updated with the incoming data stream. %
\begin{algorithm}
\small
\KwData{Multivariate signal $\By = (\by{1}, \ldots ,\by{M})\in \RR^{M\times N}$.\\
 \hspace{.75cm} Regularization parameter $\lambda>0$.\\
 \hspace{.75cm} Starting location $k_0=1$.}
 \vspace{.1cm}
\While{$k_0 < N$}{
Set $k\leftarrow k_0$\\
Initialize primal/dual bounds\\
\While{Rule~1 is satisfied}{
Set $k\leftarrow k+1$\\
\For{$m\leftarrow 1$ \KwTo $M$}{\nllabel{forins}
Update primal/dual bounds\\
\If{Rule~2 is not satisfied}{
Revise the update of primal/dual bounds
}
}
}
Estimate the change point $k_{\mathrm{rupt}}$\\
Estimate $(\widehat{\rm x}_j)_{k_0\leq j\leq k_\mathrm{rupt}}$\\
Set $k_0\leftarrow k_{\mathrm{rupt}}+1$\\
}
\KwResult{Solution $\mathbf{\widehat{x}}_{\textrm{approx}}$}
\caption{On-the-fly scheme for multivariate TV\label{algo:scheme}}
\end{algorithm}

The design of Algorithm~\ref{algo:scheme} results in specifying Rule~1 and Rule~2 allowing respectively to detect a change point and to find suitable change-point locations according to Proposition~\ref{th:FOC2}.
\subsection{Ideal case with $\HBz$ known}
\label{subsec:zKnown}

\subsubsection{Lower and upper bounds}\quad According to Proposition~\ref{th:FOC2}, the solution of the primal problem, the solution of the dual problem and the auxiliary variable have to satisfy, for every $k\in\{0,\ldots,N-1\}$,
\begin{equation}
\label{eq:const40}
\begin{cases} 
\HBu_{k+1} =  \By_{k+1}  + \HBu_{k} - \HBx_{k+1},\\
\mathrm{abs}(\HBu_{k+1}) \leq \HBz_{k+1},\\
\|\HBz_{k+1}\| = \lambda.\end{cases}
\end{equation}
with $\HBu_{0} =\HBu_{N}  =0$.
Considering the two first conditions, the prolongation condition $\HBx_{k+1} = \HBx_{k}$ leads to
\begin{equation}
\begin{cases} 
\By_{k+1} \geq \HBx_{k} - \HBz_{k+1} - \HBu_{k},\\
\By_{k+1} \leq \HBx_{k} + \HBz_{k+1} - \HBu_{k}.
\end{cases}
\label{eq:adm2Bi}
\end{equation}
Following the solution proposed for the univariate case derived in \cite{Condat_L_2013_j-ieee-spl_direct_a1D}, one can check that~\eqref{eq:adm2Bi} is satisfied by reasoning on lower and upper bounds of $\HBu_{k}$ and $\HBx_{k}$.  For every $k\in \{1,\ldots,N-1\}$, we define the lower and upper bounds of $\HBx_{k}$, labeled $\underline{\Bx}_{k}$ and $\overline{\Bx}_{k}$ respectively, as: 
\begin{equation}
\label{eq:cond0a}
\underline{\Bx}_{k} \leq \HBx_{k} \leq \overline{\Bx}_{k},
\end{equation}
and we set $\underline{\Bu}_{k}$ and $\overline{\Bu}_{k}$ as follows
\begin{equation}
\label{eq:cond0b}
\left(\forall m\in\{1,\ldots,M\}\right)\quad
 \begin{cases}
\widehat{u}_{m,k}=\underline{u}_{m,k} \quad \mbox{if} \quad \widehat{x}_{m,k}=\underline{x}_{m,k},\\
\widehat{u}_{m,k}=\overline{u}_{m,k} \quad \mbox{if} \quad \widehat{x}_{m,k}=\overline{x}_{m,k},
\end{cases}
\end{equation}
where $\underline{\Bu}_{k}$ and $\overline{\Bu}_{k}$ appear to be the upper and lower bounds of $\HBx_{k}$ respectively, i.e.
\begin{equation}
\label{eq:inequaltyDual}
\overline{\Bu}_{k} \leq \HBu_{k} \leq \underline{\Bu}_{k},
\end{equation}
as detailed in Appendix~\ref{ap:ulb}.
\vspace{0.3cm}
\subsubsection{Updating rules \& Rule~1} The prolongation condition $\HBx_{k+1} = \HBx_{k}$, which has led to \eqref{eq:adm2Bi}, becomes
\begin{equation}
\begin{cases} 
\By_{k+1} \geq \underline{\Bx}_{k} - \HBz_{k+1} - \underline{\Bu}_{k},\\
\By_{k+1} \leq \overline{\Bx}_{k} + \HBz_{k+1} - \overline{\Bu}_{k}.
\end{cases}
\label{eq:adm2Bii}
\end{equation}
If the latter condition, labeled as Rule~1, holds, then according to the primal-dual relation, we perform the update of the lower and upper bounds at location $k+1$ as follows:
\begin{equation}
\begin{cases}\label{eq:majUBi}
\underline{\Bu}_{k+1} =    \By_{k+1}  +  \underline{\Bu}_{k} - \underline{\Bx}_{k} ,\\
\overline{\Bu}_{k+1}   =    \By_{k+1} + \overline{\Bu}_{k} - \overline{\Bx}_{k},
\end{cases}
\end{equation}
and
\begin{equation}
\begin{cases}\label{eq:majXBi}
\underline{\Bx}_{k+1} =  \underline{\Bx}_{k} ,\\
\overline{\Bx}_{k+1}   = \overline{\Bx}_{k}.
\end{cases}
\end{equation}
\vskip -.3cm
\begin{remark}
Equivalently, one can systematically update primal (resp. dual) bounds according to \eqref{eq:majUBi} (resp. \eqref{eq:majXBi}) and verify that the following rewriting of the prolongation condition \eqref{eq:adm2Bii} holds:
\begin{equation}
\label{eq:condProlongDual2}
\begin{cases}
 \underline{\Bu}_{k+1} \geq -\HBz_{k+1},\\
 \overline{\Bu}_{k+1} \leq +\HBz_{k+1}.
\end{cases}
\end{equation}
\end{remark}

\vspace{.3cm}
\subsubsection{Signal prolongation \& Rule~2}\label{subsec:zKnown_signalProlongation}\quad If Rule~1 (i.e. Condition~\eqref{eq:adm2Bii} or equivalently \eqref{eq:condProlongDual2}) holds, then the assumption $\HBx_{k+1} = \HBx_{k}$ is valid. However, the upper and lower bounds may have to be updated in order to be consistent with  $\HBu_{k+1}\in[-\HBz_{k+1},+\HBz_{k+1}]$. According to \eqref{eq:inequaltyDual}, this condition requires to verify that the following Rule~2 holds:
\begin{equation}
\begin{cases}
 \underline{\Bu}_{k+1} \leq +\HBz_{k+1},\\
 \overline{\Bu}_{k+1} \geq -\HBz_{k+1}.
\end{cases}
\label{eq:zcons}
\end{equation}
For every $m\in\{1,\ldots,M\}$, three configurations can be encountered:
\begin{itemize}
\item When both Conditions \eqref{eq:zcons} are satisfied, the bounds are left unchanged.
\item When $\underline{u}_{m,k+1} = \underline{u}_{m,k}  + y_{m,k+1} - \underline{x}_{m,k} > +\widehat{\constr}_{m,k+1}$,
then the updating rules specified in \eqref{eq:majXBi} have under-evaluated the bound
\begin{equation}
\underline{\nu}_m \equiv \underline{x}_{m,j}\quad\left(\forall j\in\{k_0,\ldots,k+1\}\right)
\end{equation}
where $k_0$ denotes the last starting location of a new segment. Since $ \underline{u}_{m,k+1}$ is upper-bounded by $+\widehat{\constr}_{m,k+1}$ and, that for such a value it can be shown (see Appendix~\ref{app:proofXmin}) that
\begin{equation}
\label{eq:updateXmin}
\underline{\nu}_m =  \underline{x}_{m,k} + \frac{\underline{u}_{m,k+1} - \widehat{z}_{m,k+1} }{k - k_0+1}, 
\end{equation}
we propose the following updates
\begin{equation}
\label{eq:updateXlowbnd}
\begin{cases}
 (\forall j\in\{k_0,\ldots,k+1\})\quad \underline{x}_{m,j} = \underline{\nu}_m,\\
 \underline{u}_{m,k+1}=+\widehat{\constr}_{m,k+1}.
 \end{cases}
\end{equation}
\item When $\overline{u}_{m,k+1}< - \widehat{\constr}_{m,k+1}$, then it results that the upper bound
\begin{equation}
\overline{\nu}_m\equiv \overline{x}_{m,j}\quad\left(\forall j\in\{k_0,\ldots,k+1\}\right)
\end{equation}
has been over-evaluated. Similarly, since $\overline{u}_{m,k+1}$ is lower bounded by $-\widehat{\constr}_{m,k+1}$, we can show that the upper bound
\begin{equation}
\overline{\nu}_m =  \overline{x}_{m,k} + \frac{\overline{u}_{m,k+1} + \widehat{z}_{m,k+1} }{k - k_0+1},
\end{equation}
permits to ensure the consistency of the following updates
\begin{equation}
\label{eq:updateXuppbnd}
\begin{cases}
 (\forall j\in\{k_0,\ldots,k+1\})\quad \overline{x}_{m,j} = \overline{\nu}_m,\\
 \overline{u}_{m,k+1}=-\widehat{\constr}_{m,k+1}.
 \end{cases}
\end{equation}
\end{itemize}
\vspace{0.3cm}

\subsubsection{Estimate of the change point $k_{\mathrm{rupt}}$}\label{eq:createcp}\quad When Rule 1 does not hold, a change point has to be created. For every $m\in\{1,\ldots,M\}$, we can distinguish three cases:
\begin{itemize}
\item When $\underline{u}_{m,k+1}=\underline{u}_{m,k} + y_{m,k+1} - \underline{x}_{m,k} < - \widehat{\constr}_{m,k+1}$, then, since $\underline{u}_{m,k}$ is bounded, it means that $\underline{x}_{m,k}$ is over-evaluated and therefore a negative amplitude change has to be introduced on the $m$-th component in the time index set $\{k_0,\ldots,k\}$ in order to decrease its value. Following Proposition~\ref{th:FOC2} and Eq.~\eqref{eq:inequaltyDual}, the set of locations $\kappa_m$ suitable for a change-point on the $m$-th component reads:
\begin{equation}
\kappa_m = \{ j\in\{k_0,\ldots,k\}\quad|\quad \underline{u}_{m,j} = + \widehat{\constr}_{m,j}\}.
\end{equation} 
Such locations correspond to the indexes where the value of the bound $\underline{u}_{m,j}$ has been updated in order to be consistent with the condition $\widehat{u}_{m,j}\in[-\widehat{\constr}_{m,j}, \widehat{\constr}_{m,j}]$ (see the previous paragraph)
\item When $\overline{u}_{m,k+1}>+\widehat{\constr}_{m,k+1}$, then a positive amplitude change has to be introduced in the $m$-th component within the time index $\{k_0,\ldots,k\}$. The set of locations suitable for a change-point on the $m$-th component reads:
\begin{equation}
\kappa_m = \{ j\in\{k_0,\ldots,k\}\quad|\quad \overline{u}_{m,j} = - \widehat{\constr}_{m,j}\}.
\end{equation}
This set of locations corresponds to indexes where the value of the bound $\overline{u}_{m,j}$ was updated in order to be consistent with $\widehat{u}_{m,j}\in[-\widehat{\constr}_{m,j}, \widehat{\constr}_{m,j}]$.
\item Else, when component $m$ does satisfy \eqref{eq:adm2Bi}, then we set $\kappa_m =\{k_0,\ldots,k\}$.
\end{itemize}

The change-point location $k_\mathrm{rupt}$ corresponds to the last location suitable for introducing the adequate amplitude change on each component, i.e.,
\begin{equation}
\label{eq:changePointLocation}
k_\mathrm{rupt} = \max_{j\in\cap_{m=1}^M \kappa_m} j.
\end{equation}

Once the change point location has been specified, we are able to assign a value to $\left(\HBx_j\right)_{k_0\leq j\leq k_{\mathrm{rupt}}}$. When a negative amplitude change is detected on the $m$-th component, we set
\begin{equation}
\label{eq:negAmpChange}
(\forall j\in\{k_0,\ldots,k_\mathrm{rupt}\})\quad \widehat{x}_{m,j} = \underline{x}_{m,k+1},
\end{equation}
in consistence with \eqref{eq:cond0b}. Similarly, when a positive amplitude change is detected, we set
\begin{equation}
\label{eq:posAmpChange}
(\forall j\in\{k_0,\ldots,k_\mathrm{rupt}\})\quad \widehat{x}_{m,j} = \overline{x}_{m,k+1}.
\end{equation}
\vspace{0.1cm}
\subsubsection{Starting a new segment}\quad When a segment has been created, we start the detection of a new segment considering $k_0=k_\mathrm{rupt}+1$ as long as $k_0 < N$. 

According to \eqref{eq:optPrimalDual} and by definition of the bounds, for every $k\in\{1,\dots, N\}$
\begin{equation}
\label{eq:pdr}
\begin{cases}
\underline{\Bx}_{k} = {\By}_{k} - \underline{\Bu}_{k}  + \HBu_{k-1},\\
\overline{\Bx}_{k} = {\By}_{k} - \overline{\Bu}_{k}  + \HBu_{k-1}.
\end{cases}
\end{equation}
In particular, for $k=k_0$, combining \eqref{eq:2DOC}, \eqref{eq:cond0a}, \eqref{eq:cond0b} and \eqref{eq:majUBi} allows us to find the following initialization procedure
\begin{equation}
\label{eq:initZknown}
\begin{array}{ll}
\begin{cases}
\underline{\Bu}_{k_0} = +\HBz_{k_0}, \\
\overline{\Bu}_{k_0} = -\HBz_{k_0},
\end{cases}\\
\begin{cases}
\underline{\Bx}_{k_0} = \By_{k_0} - \HBz_{k_0} + \HBu_{k_0-1},\\
\overline{\Bx}_{k_0} = \By_{k_0} + \HBz_{k_0} + \HBu_{k_0-1},\\
\end{cases}
\end{array}
\end{equation}
where the value of $\HBu_{k_0-1}$ is given according to Proposition~\ref{th:FOC2}. In addition, according to the writing of \eqref{eq:const40}, $\HBu_{0} = 0$.
\vskip \baselineskip

\subsection{Estimation of the auxiliary multivariate vector $\HBz$}

In order to describe the generic behavior of the multivariate on-the-fly algorithm, we have so far assumed $\HBz$ to be known a priori. We now focus on the simultaneous estimation of the multivariate vector $\HBz$ and of the multivariate signal $\HBx$. 

To provide an on-the-fly approximate solution, we propose:
\begin{itemize}
 \item to build a piece-wise constant estimator $\widetilde{\mathbf{z}}$ of $\HBz$,
 \item to only consider amplitude changes jointly on all components $m\in\{1,\ldots,M\}$. 
\end{itemize}
\vspace{0.3cm}

\subsubsection{Piece-wise constant estimator of $\HBz$}\quad The proposed estimate is assumed to be constant between each change-point with values belonging to the predefined set $\mathcal{Q}$ defined in Section~\ref{ssec:approx}. For each candidate value $\zeta^{(q)}$ with $q\in\{1,\ldots,|\mathcal{Q}|\}$, we create upper and lower bounds labeled $\underline{\Bu}_{k}^{(q)}$, $\overline{\Bu}_{k}^{(q)}$, $\underline{\Bx}_{k}^{(q)}$, and $\overline{\Bx}_{k}^{(q)}$. They are initialized at each new segment location $k_0$ and are updated independently according to \eqref{eq:majUBi} and \eqref{eq:majXBi} until the prolongation condition
 \begin{equation}
\label{eq:condProlongDual2_v2}
\begin{cases}
 \underline{\Bu}_{k+1}^{(q)} \geq -\zeta^{(q)},\\
 \overline{\Bu}_{k+1}^{(q)} \leq +\zeta^{(q)},
\end{cases}
\end{equation}
based on \eqref{eq:condProlongDual2}, does not hold anymore.
In the following, we investigate how to modify the algorithm described in Section~\ref{subsec:zKnown}, to account for the automated selection of $\mathbf{\widetilde{\constr}}$ in $\mathcal{Q}$. The resulting algorithm is reported in Algorithm~\ref{algo:onlineBiTV}.
\vspace{0.3cm}

\subsubsection{Estimate of the change point $k_\mathrm{rupt}^{(q)}$\label{sec:zEst_changepoint}}
\begin{figure}[t]
\centering \hskip .3cm
\includegraphics[scale=.45]{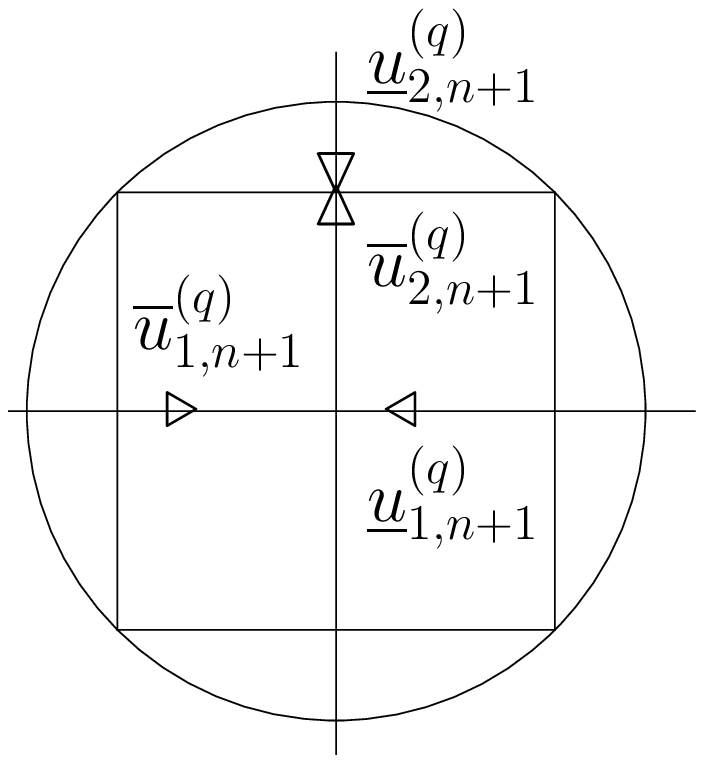}\qquad
\includegraphics[scale=.45]{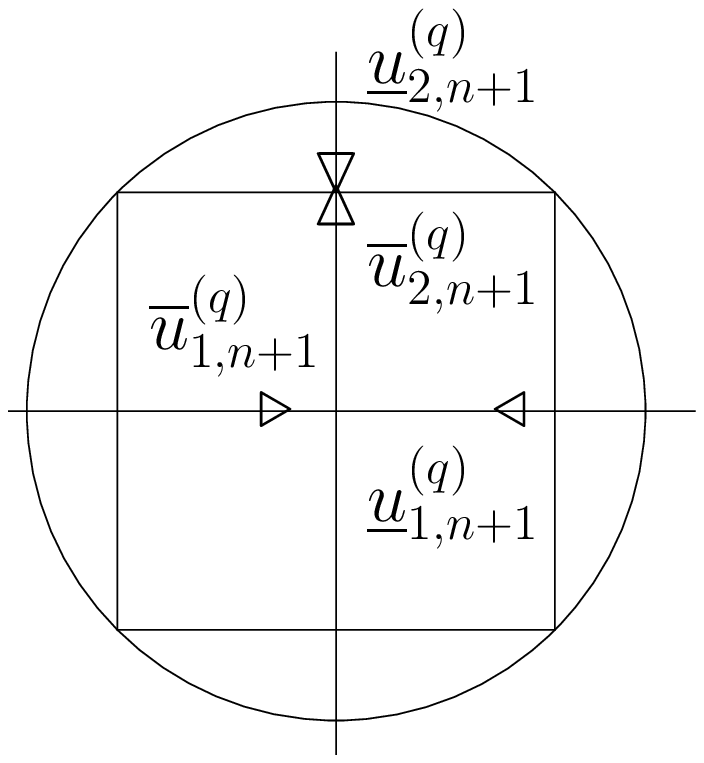}\
\caption{\textbf{Example of configurations leading to the detection of a change-point}. In this example $M=2$, $\zeta^{(q)}_1=\zeta^{(q)}_2=\lambda/\sqrt{2}$. Since $\underline{u}^{(q)}_{2,k+1}>\zeta_2^{(q)}$, condition \eqref{eq:condProlongDual2_v2} is violated. The left (resp. right) plot displays the configuration $\overline{u}_{1,k+1}^{(q)}+\underline{u}_{1,k+1}^{(q)}<0$ (resp. $\overline{u}_{1,k+1}^{(q)}+\underline{u}_{1,k+1}^{(q)}\geq0$) described in Section~\ref{sec:zEst_changepoint}.\label{fig:condCreationChangePoint}}
\end{figure}

\quad For every $ q\in \{1,\ldots,|\mathcal{Q}|\}$, we create change points as described in Section~\ref{eq:createcp}. The main difference consists in the restriction to simultaneous change points. As detailed after Proposition~\ref{th:FOC}, non-simultaneous changes have a zero probability to occur. The restriction to simultaneous change-points will thus not impact the solution. It results that if there exists at least one component $m\in\{1,\ldots,M\}$ such that $\underline{u}_{m,k+1}^{(q)} < -\zeta_m^{(q)}$ (resp. $\overline{u}_{m,k+1}^{(q)} > \zeta_m^{(q)}$), then
\begin{equation}
\kappa_m^{(q)}=\{j\in\{k_0,\ldots,k\}\;|\;\underline{u}_{m,j}^{(q)}=+\zeta_m^{(q)}\}\\
\end{equation}
\begin{equation}(\mbox{resp.} \quad  \kappa_m^{(q)}=\{j\in\{k_0,\ldots,k\}\;|\;\overline{u}_{m,j}^{(q)}=-\zeta_m^{(q)}\}),\end{equation}
 and, $\forall m_-\neq m$ such that $\underline{u}_{m_-,k+1}^{(q)} + \overline{u}_{m_-,k+1}^{(q)} <0$,  then
\begin{equation}
\kappa_{m_-}^{(q)}=\{j\in\{k_0,\ldots,k\}\;|\;\underline{u}_{m_-,j}^{(q)}=+\zeta_m^{(q)}\}\\
\end{equation}
or, $\forall m_+\neq m$ such that  $\underline{u}_{m_+,k+1}^{(q)} + \overline{u}_{m_+,k+1}^{(q)} \geq 0$, then
\begin{equation}
\kappa_{m_+}^{(q)}=\{j\in\{k_0,\ldots,k\}\;|\;\underline{u}_{m_+,j}^{(q)}=+\zeta_m^{(q)}\}.
\end{equation}
A bivariate example of these configurations where the second component breaks Condition~$\eqref{eq:condProlongDual2_v2}$ is provided in Fig.~\ref{fig:condCreationChangePoint}. The change-point location $k_\mathrm{rupt}^{(q)}$ and the assignment of $\HBx^{(q)}$ on the current segment follow \eqref{eq:changePointLocation}, \eqref{eq:negAmpChange} and \eqref{eq:posAmpChange}.
\vspace{0.3cm}

\subsubsection{Estimate of the change point $k_\mathrm{rupt}$\label{sec:findKrupt}} According to the previous paragraph, the piece-wise estimation procedure leads to several possible change-point locations (at most $|\mathcal{Q}|$). Here we select the solution indexed by $q^*$ with tightest bounds $\underline{\Bx}^{(q^*)}$ and $\overline{\Bx}^{(q^*)}$, i.e.,
\begin{equation}
\label{eq:epsilonChoiceZeta}
q^* \in \underset{1\leq q \leq |\mathcal{Q}|}{\mathrm{Argmin}}\;  \left\| \left(\overline{\Bx}_{k_\mathrm{rupt}^{(q)}}^{(q)} - \underline{\Bx}_{k_\mathrm{rupt}^{(q)}}^{(q)} \right) \boldsymbol{\sigma}^{-1}\right\|^2,
\end{equation}
with 
{\begin{equation}
\boldsymbol{\sigma} = {\mathop{\rm diag}}(\sigma_1,\ldots,\sigma_M),
\end{equation}
where, for every $m\in\{1,\ldots,M\}$, $\sigma_m$ stands for the standard deviation of $\boldsymbol{y}_m$. The factor $\boldsymbol{\sigma}^{-1}$ permits to ensure that every component contributes equally to the criterion~\eqref{eq:epsilonChoiceZeta}. When the minimizer of \eqref{eq:epsilonChoiceZeta} is not unique, we select the index $q^*$ yielding the largest $k_\mathrm{rupt}^{(q^*)}$. In other words, we choose the set of auxiliary variables which permits to hold the prolongation condition \eqref{eq:condProlongDual2_v2} as long as possible. 

Therefore, it finally leads to an index $q^*$ which permits to estimate $k_{\mathrm{rupt}} =k_{\mathrm{rupt}}^{(q^ *)}$ and,  
\begin{equation}
(\forall j\in\{k_0,\ldots,k_\mathrm{rupt}\})\quad
\tilde{\Bz}_{j} = \zeta^{(q^*)},\;\HBx_{j} = \HBx_{j}^{(q^*)}. 
\label{eq:assignZandX}
\end{equation}

The starting location for the next segment is then, $k_0 = k_\mathrm{rupt}+1$, and the algorithm iterates as long as $k_0<N$.
\vspace{0.3cm}

\subsubsection{Starting a new segment}\quad Let us consider the location $k_0$ of a new segment. For every $q\in\{1,\ldots,|\mathcal{Q}|\}$, the \emph{initialization} step can be recast into
\begin{equation}
\label{eq:initZunknown}
\begin{array}{ll}
\begin{cases}
\underline{\Bu}_{k_0}^{(q)} = +\zeta^{(q)},
&\overline{\Bu}_{k_0}^{(q)} = -\zeta^{(q)},\\
\underline{\Bx}_{k_0}^{(q)} = \By_{k_0} - \zeta^{(q)} + \HBu_{k_0-1},
&\overline{\Bx}_{k_0}^{(q)} = \By_{k_0} + \zeta^{(q)} + \HBu_{k_0-1},
\end{cases}
\end{array}
\end{equation}
with $\HBu_{0} = 0$. \vspace{-.3cm}
\begin{remark} The initialization step \eqref{eq:initZunknown} implicitly depends on the estimation of $\HBz$ made on the last segment through the term $\HBu_{k_0-1}$. Simulations have shown that \eqref{eq:initZunknown} may lead to an inconsistent solution $\HBx$ as soon as $\HBz$ has been poorly estimated on a segment. Empirically, a better approximation of the iterative solution is observed if each segment is treated independently, i.e.,
\end{remark}
\begin{equation}
\label{eq:initZunknownBis}
\begin{array}{ll}
\begin{cases}
\underline{\Bu}_{k_0}^{(q)} = +\zeta^{(q)},
&\overline{\Bu}_{k_0}^{(q)} = -\zeta^{(q)},\\
\underline{\Bx}_{k_0}^{(q)} = \By_{k_0} - \zeta^{(q)},
&\overline{\Bx}_{k_0}^{(q)} = \By_{k_0} + \zeta^{(q)}.
\end{cases}
\end{array}
\end{equation}

\begin{algorithm}[!t]
\small
\KwData{Multivariate signal $\By = (\by{1}, \ldots ,\by{M})\in \RR^{M\times N}$.\\
 \hspace{.75cm} Regularization parameter $\lambda>0$.\\
 \hspace{.75cm} Predefined set $\mathcal{Q}=\{\zeta^{(1)},\ldots,\zeta^{(|\mathcal{Q}|)}\}$.\\
 \hspace{.75cm} Starting location $k_0=1$.}
 \vspace{.1cm}
\While{$k_0 < N$}{
\For{$q\leftarrow 1$ \KwTo $|\mathcal{Q}|$}{\nllabel{forins}
Set $k\leftarrow k_0$\\
Initialize primal/dual bounds according to \eqref{eq:initZunknownBis}\\
\While{\eqref{eq:condProlongDual2_v2} is satisfied}{
Set $k\leftarrow k+1$\\
\For{$m\leftarrow 1$ \KwTo $M$}{\nllabel{forins}
Update primal/dual bounds\\
\If{$\underline{u}_{m,k+1}^{(q)}  > +\zeta_m^{(q)}$ or $\overline{u}_{m,k+1}^{(q)} < - \zeta_m^{(q)}$}{
Revise the update of primal/dual bounds
}
}
}
Estimate $k_\mathrm{rupt}^{(q)}$ and $(\HBx_j^{(q)})_{k_0\leq j \leq k_\mathrm{rupt}^{(q)}}$ according to Section~\eqref{sec:zEst_changepoint}\\
}
Estimate $k_\mathrm{rupt}\in (k_\mathrm{rupt}^{(q)})_{1\leq q \leq | \mathcal{Q}|}$ according to Section~\ref{sec:findKrupt}\\
Estimate $(\HBx_j)_{k_0\leq j\leq k_\mathrm{rupt}}$ and $(\tilde{\Bz}_j)_{k_0\leq j\leq k_\mathrm{rupt}}$ according to \eqref{eq:assignZandX}\\
Set $k_0\leftarrow k_{\mathrm{rupt}}+1$\\	
}

\KwResult{Solution $\HBx$}
\caption{\label{algo:onlineBiTV}On-the-fly Multivariate TV}
\end{algorithm}

\section{Performance assessment\label{sec:res}}
\subsection{Experimental setting}

Unless specified otherwise, we consider that data consist of a $M$-multivariate piece-wise constant signal $\Bx\in\mathbb{R}^N$ (solid black), to which a centered Gaussian noise $\boldsymbol{\epsilon}$ is additively superimposed: $\By=\Bx + \boldsymbol{\epsilon}\in\mathbb{R}^{M \times N}$.

Signal $\Bx$ is generated as follows. 
First the length of each segment is drawn according to a folded Gaussian distribution $\mathcal{N}(12.5,16.25)$. 
Then, for each $m\in\{1,\ldots,M\}$, the amplitudes of the corresponding changes are drawn independently from a Gaussian distribution $\mathcal{N}(2,0.4)$.

The exact minimizer of \eqref{eq:pbPrimal1}, labeled $\HBx$, is computed by means of the ADMM algorithm proposed in \cite{Wahlberg_B_2012_ifac_admmactvrep}. Iterations are stopped when the relative criterion error is lower than $10^{-10}$.
The proposed solution computed with the predefined set $\mathcal{Q}$ is denoted $\HBx_{\textrm{approx},\mathcal{Q}}$.

In a second set of experiments (see \ref{sec:onlineperf}), the proposed on-the-fly algorithmic solution will be compared to an on-the-fly ADMM solution.

\subsection{Design of $\mathcal{Q}$\label{sec:designQ}}
We propose to compare solutions $\HBx_{\textrm{approx},\mathcal{Q}}$ obtained with two sets $\mathcal{Q}=\{\zeta^{(1)},\ldots, \zeta^{(|\mathcal{Q}|)}\}$ in the bivariate case (i.e., $M=2$) for $N=10^4$. For both configurations, we choose
\begin{equation}
(\forall q\in\{1,\ldots,|\mathcal{Q}|\})\quad \zeta^{(q)}=\left(\lambda\cos(\theta_q), \lambda\sin(\theta_q)\right)
\end{equation}
with $\quad\theta_q\in[0,\pi/2]$. The first solution consists to homogeneously cover the $\ell_2$ ball such that, for some some parameter $R\in \mathbb{N}_*$, 
$\theta_q=q\pi/2^{R+1}$ and $|\mathcal{Q}|= \sum_{q'=0}^{R-1} 2^{q'}$.
The second solution draws a set of the same size whose values $(\theta_q)_{1\leq q \leq  |\mathcal{Q}|}$ follow a uniform distribution on $[0,\pi/2]$.

Two experimental settings are investigated. In the first one, $\by{1}$ is one order of magnitude larger than $\by{2}$ (Fig.~\ref{fig:designQexp1}, left plots) whereas in the second one, both are of the same order of magnitude (Fig.~\ref{fig:designQexp1}, right plots).

\begin{figure}[t]
\centering \includegraphics[scale=.28, clip=true, trim=-.3cm 0cm 0cm 0cm]{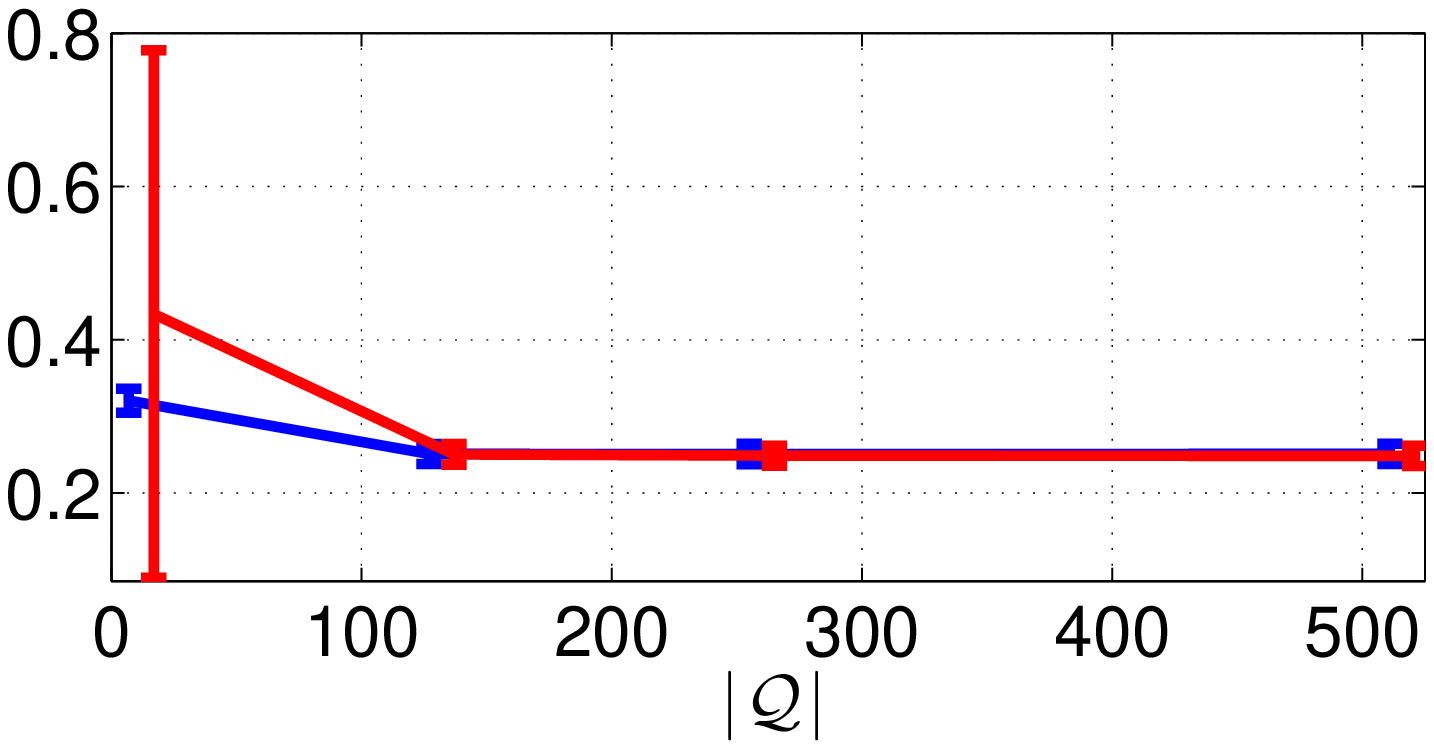}
\includegraphics[scale=.29, clip=true, trim=-.8cm 0cm 0cm 0cm]{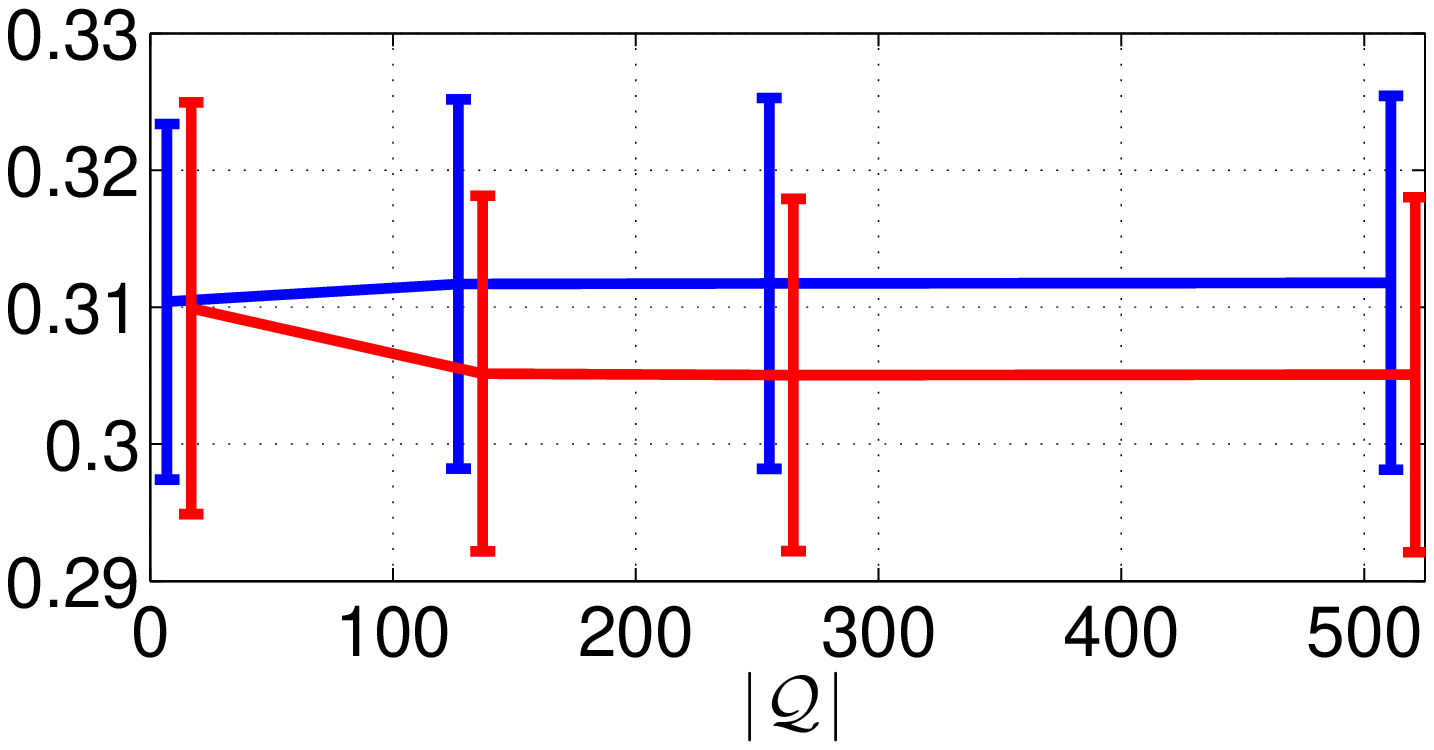}
\vskip .1cm
\includegraphics[scale=.28, clip=true, trim=.85cm 1.4cm 1.7cm .9cm]{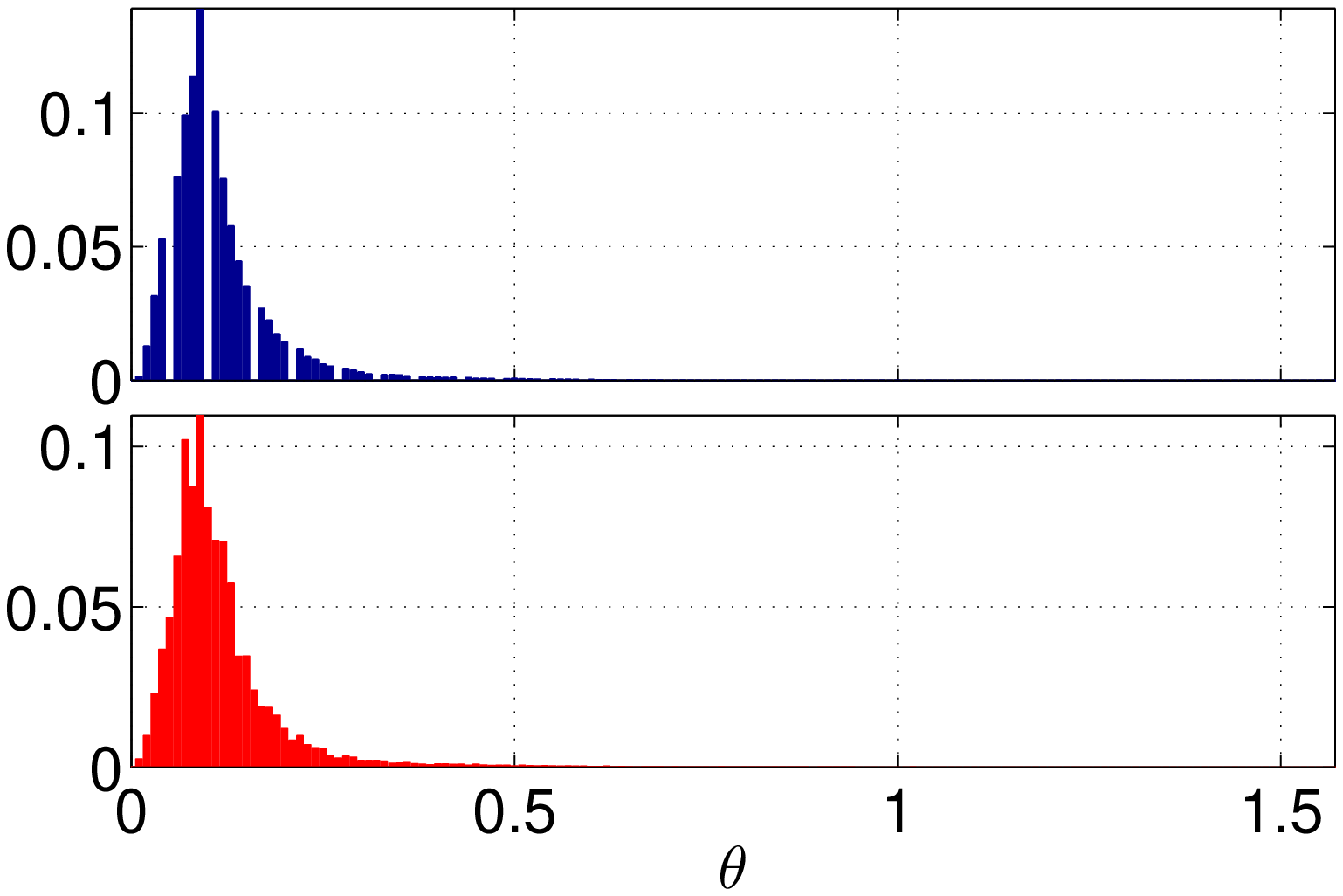}
\includegraphics[scale=.28, clip=true, trim=.85cm 1.4cm 1.7cm .9cm]{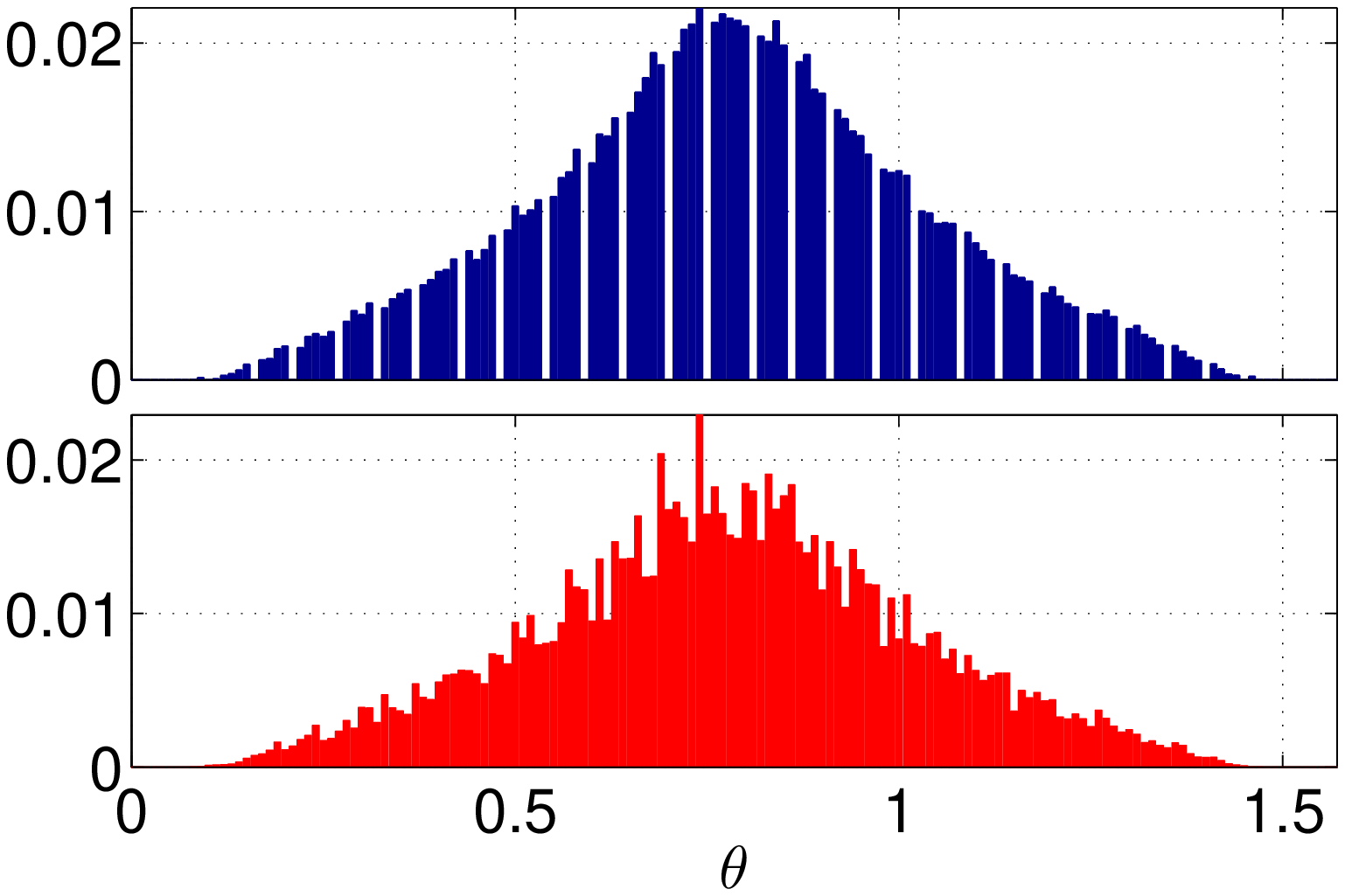}
\caption{\textbf{Influence of the design of $\mathcal{Q}$.} Comparison over 100 realizations of an homogeneous covering of the $\ell_2$-ball (blue) against a random covering (red). Two experimental settings are considered depending on whether if $\by{1}$ is one order of magnitude larger than $\by{2}$ (left) or not (right). Top:  $\mathrm{MSE}(\HBx_{\textrm{approx},\mathcal{Q}},\HBx)$ for different set sizes $|\mathcal{Q}|$. Bottom (2nd and 3rd lines): distributions of $\theta_{q^*}$ where $q^*$ has been selected by criterion \eqref{eq:epsilonChoiceZeta} for $|\mathcal{Q}|=127$.\label{fig:designQexp1}}
\end{figure}

Estimation performances in terms of mean squared error $\mathrm{MSE}(\HBx_{\textrm{approx},\mathcal{Q}},\HBx)=\widehat{\mathbb{E}}[\frac{1}{N}\| \HBx_{\textrm{approx},\mathcal{Q}} - \HBx\|^2]$ (where $\widehat{\mathbb{E}}$ stands for the sample mean estimator computed over 100 realizations) are reported on the first line. It shows that a random covering of the $\ell_2$-ball provides solutions as good as the homogeneous covering up to the limit of $|\mathcal{Q}|$ small.

On the 2nd and 3rd lines, the distributions of $\theta_{q^*}$, where $q^*$ has been selected by criterion \eqref{eq:epsilonChoiceZeta}, are reported for $|\mathcal{Q}|=127$. These histograms show the impact of the relative amplitude of the components on the distribution $\theta_{q^*}$: components with same order of magnitude yield a symmetric distribution while unbalanced components yield an asymmetric distribution. For instance, in Fig.~\ref{fig:designQexp1} (right plots), it appears more meaningful to draw $\theta_q$ according to a Gaussian distribution than to a uniform distribution. Therefore, if one has some knowledge of components amplitudes, this can be incorporated to better design the set $\mathcal{Q}$. This will also decrease the computational cost discussed in section \ref{sec:onlineperf}.

In the following, we restrict ourselves to a random covering of the $\ell_2$ ball.

\subsection{Offline performance}
In this section, we focus on the comparison of offline performance, extended for $M=10$, for two different signal-to-noise ratios (SNRs), namely $4$dB and $10$dB.\\

\noindent\textbf{Qualitative impact of $|\mathcal{Q}|$ on $\HBx_{\textrm{approx},\mathcal{Q}}$.}~ For a single realization of noise, $\HBx_{\textrm{approx},\mathcal{Q}}$ and $\mathbf{\widehat{x}}$ are plotted Fig.~\ref{fig:solution} for $\lambda=29$, adjusted to provide the best visual (qualitative) performance. Solution $\HBx_{\textrm{approx},\mathcal{Q}}$ for $|\mathcal{Q}|=5\times10^4$ (light orange) provides a visually better approximation of $\mathbf{\widehat{x}}$ (dashed blue) than for $|\mathcal{Q}|=10^3$ (mixed red).\\

\noindent\textbf{Estimation performance $\HBx_{\textrm{approx},\mathcal{Q}}$ vs. $\HBx$.} The quality of the approximation is further quantified Fig.~\ref{fig:approxQuality} in terms of $\mathrm{MSE}(\HBx_{\textrm{approx},\mathcal{Q}},\HBx)$ as a function of $\lambda$ for different $|\mathcal{Q}|$.

It shows that the MSE systematically decreases when $|\mathcal{Q}|$ increases. 
Further, on the examples considered here and depending on $\lambda$, using $|\mathcal{Q}| \geq 10^4$ no longer yields significantly improved solutions, thus showing that the selection of $|\mathcal{Q}|$ does not require a complicated tuning procedure.\\

\noindent\textbf{Estimation performance $\HBx$ vs. $\Bx$ and $\HBx_{\textrm{approx},Q}$ vs. $\Bx$.}~ Let us now compare the absolute quality of the solutions against $\Bx$.

$\mathrm{MSE}(\HBx,\Bx)$ and $\mathrm{MSE}(\HBx_{\textrm{approx},\mathcal{Q}},\Bx)$ for different $|\mathcal{Q}|$, are reported in Fig.~\ref{fig:estimPerf}. 
MSEs are consistent with the previous paragraph: it shows that increasing $|\mathcal{Q}|$ up to a certain value permits to significantly lower the MSE. However, $\HBx$ has a lower estimation error than $\HBx_{\textrm{approx},Q}$.\\

\begin{figure}[t]
\centering \hskip -.3cm\includegraphics[scale=.3, clip=true, trim=4cm 5cm .95cm 1cm]{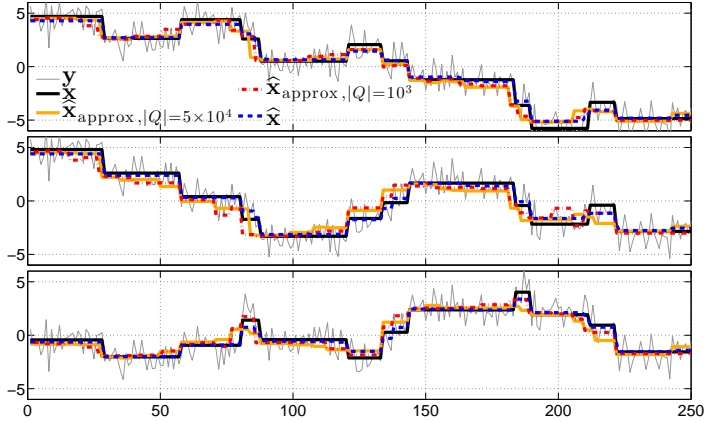}
\caption{\label{fig:solution}\textbf{Qualitative impact of $|\mathcal{Q}|$ on $\HBx_{\textrm{approx},\mathcal{Q}}$.} For visibility, only 3 components out of $M=10$ are displayed. ${\rm SNR}=4$dB. $\HBx_{\textrm{approx},\mathcal{Q}}$ for $|\mathcal{Q}|=5\times 10^4$ is more satisfying than for $|\mathcal{Q}|=10^3$ since it has more discontinuities in common with $\HBx$.}
\end{figure}

\subsection{Online performance\label{sec:onlineperf}}
In this section we focus on the comparison between two online solutions. The first one is derived from the proposed on-the-fly algorithm whereas the second one is based on an iterative algorithm.

Comparison is made for different values of $\lambda$ on a signal $\Bx\in\mathbb{R}^{M\times N}$ ($N=400$) to which a Gaussian noise is superimposed such that ${\rm SNR}=3$dB. Performance are provided for $M=2$ and $M=5$ components.\\

\noindent\textbf{Proposed online solution $\HBx_{\mathrm{online},\mathcal{Q}}$.}~ As the time step $k$ increases, $\HBx_{\textrm{approx},\mathcal{Q}}$ is only computed up to the last $k_0$ and the algorithm has not yet output a solution on $\{k_0+1,\ldots,k\}$. In that sense, the solution is said to be "on-the-fly". However, a solution $\HBx_{\mathrm{online},\mathcal{Q}}$, providing an online approximation of $\Bx$, can be output up to $k$ by imposing limit conditions at $k$.\\

\noindent\textbf{Windowed iterative solution $\HBx_{\mathrm{win},K}$.}~ We consider a naive online ADMM version, where at each time step $k$ a solution $\HBx_{\mathrm{win},K}$ is computed by optimizing over the previous $K$ points. The choice of $K$ is of critical importance. On the one hand, if this value is too small, the observer may miss amplitude changes in the multivariate data stream. On the other hand, if the window size is too large, the computational cost may be too high to handle any online observation. Three window sizes have been investigated, respectively $K=20$, $50$ and $80$.\\

\noindent\textbf{Computational cost.}~
Comparisons of median computational costs per incoming sample (in seconds), over 10 realizations of noise, are reported Fig.~\ref{fig:onlinePerf} (left plots) as functions of $\lambda$.

As expected, we observe that the computational cost does increase along with the size of $\mathcal{Q}$. Therefore, $|\mathcal{Q}|$ acts as a trade-off between the computational cost and the MSE. However, the computational cost of $\HBx_{\mathrm{online},\mathcal{Q}}$ is still several orders of magnitude lower than the one associated to the online ADMM. Interestingly, computational costs are comparable for $M=2$ (top left plot) and $M=5$ (bottom left plot). Note that a warm-up starting strategy for online ADMM only reduces by a factor two the computational cost with respect to the implementation displayed in  Fig.~\ref{fig:onlinePerf}.

The computational cost of $\HBx_{\mathrm{online},\mathcal{Q}}$ could still be reduced in two ways. First, one could design the set $\mathcal{Q}$ according to a priori knowledge of components amplitudes (see \ref{sec:designQ}). Second, one could also benefit from the separable form of the algorithm and compute solutions $\HBx^{(q)}$ in parallel for every $q\in\{1,\ldots,|\mathcal{Q}|\}$.\\

\noindent\textbf{Change-point detection accuracy.}~ The Jaccard index $J(\boldsymbol{\alpha},\boldsymbol{\beta})\in[0,1]$ between any $\boldsymbol{\alpha}$ and $\boldsymbol{\beta}\in[0,1]^N$ is defined as \cite{Jaccard_P_1901_bsvsn_dfabdrv,Hamon_R_2014_arxiv_hscbspsg}
\begin{equation}
J(\boldsymbol{\alpha},\boldsymbol{\beta}) = \frac{\sum_{i=1}^N \min(\alpha_i, \beta_i)}{
\sum_{\substack{
   1 \leq i \leq N \\
   \alpha_i>0,\beta_i>0
  }}^N \frac{\alpha_i + \beta_i}{2} + \sum_{\substack{
   1 \leq i \leq N \\
   \beta_i=0
  }} \alpha_i 
  +  \sum_{\substack{
   1 \leq i \leq N \\
   \alpha_i=0
  }} \beta_i 
  }.
  \end{equation}
It varies from $0$, when $\boldsymbol{\alpha}\cap\boldsymbol{\beta}=\emptyset$, up to $1$ when $\boldsymbol{\alpha}=\boldsymbol{\beta}$. The Jaccard index is a demanding measure: As an example, if $\boldsymbol{\beta}\in\{0,1\}^N$ is the truth and if $\boldsymbol{\alpha}\in\{0,1\}^N$ has correctly  identified half non-zero values of $\boldsymbol{\beta}$ but has misidentified the other half, then $J(\boldsymbol{\alpha},\boldsymbol{\beta})=1/3$.

The Jaccard index is used to measure the similarity between change-point locations of $\Bx$ and those obtained during the computation of $\HBx_{\mathrm{win},K}$ and $\HBx_{\mathrm{online},\mathcal{Q}}$. To this end, we consider the change-point indicator vector $\boldsymbol{r}=(r_i)_{1\leq i \leq N}$ of $\Bx$ (as well as $\widehat{\boldsymbol{r}}_{\mathrm{win},K}$ and $\widehat{\boldsymbol{r}}_{\mathrm{online},\mathcal{Q}}$ respectively associated to $\HBx_{\mathrm{win},K}$ and $\HBx_{\mathrm{online},\mathcal{Q}}$), defined as
\begin{equation}
r_{i} =\left\{
              \begin{array}{ll}
                1, & \hbox{if $\Bx$ has a change-point at location $i$,} \\
                0, & \hbox{otherwise.}
              \end{array}
            \right.
\end{equation}
In order to incorporate a tolerance level on change-point locations, $\boldsymbol{r}$, $\widehat{\boldsymbol{r}}_{\mathrm{win},K}$ and $\widehat{\boldsymbol{r}}_{\mathrm{online},\mathcal{Q}}$ are first convolved with a Gaussian kernel of size $10$ with a standard deviation of $3$.

$J(\widehat{\boldsymbol{r}}_{\mathrm{win},K},\boldsymbol{r})$ and $J(\widehat{\boldsymbol{r}}_{\mathrm{online},\mathcal{Q}},\boldsymbol{r})$ are averaged over 10 realizations of noise and reported in Fig.~\ref{fig:onlinePerf} (right plots) as functions of $\lambda$ for \mbox{different set size $|\mathcal{Q}|$ and window size $K$.}

Performance show that $J(\widehat{\boldsymbol{r}}_{\mathrm{online},\mathcal{Q}},\boldsymbol{r})\geq J(\widehat{\boldsymbol{r}}_{\mathrm{win},K},\boldsymbol{r})$ for almost all $\lambda$ and $\vert \mathcal{Q}\vert$. Therefore, $\HBx_{\mathrm{online},\mathcal{Q}}$ provides a better online detection of change-points of $\Bx$. It also show that $J(\widehat{\boldsymbol{r}}_{\mathrm{online},\mathcal{Q}},\boldsymbol{r})$ does not vary significantly with $|\mathcal{Q}|$ but slightly decreases with $M$. Indeed, as $M$ increases, the prolongation condition \eqref{eq:condProlongDual2_v2} is more likely to be violated, thus leading to more change-points. 

\begin{figure}[t]
\vskip .1cm
\centering\includegraphics[scale=.32]{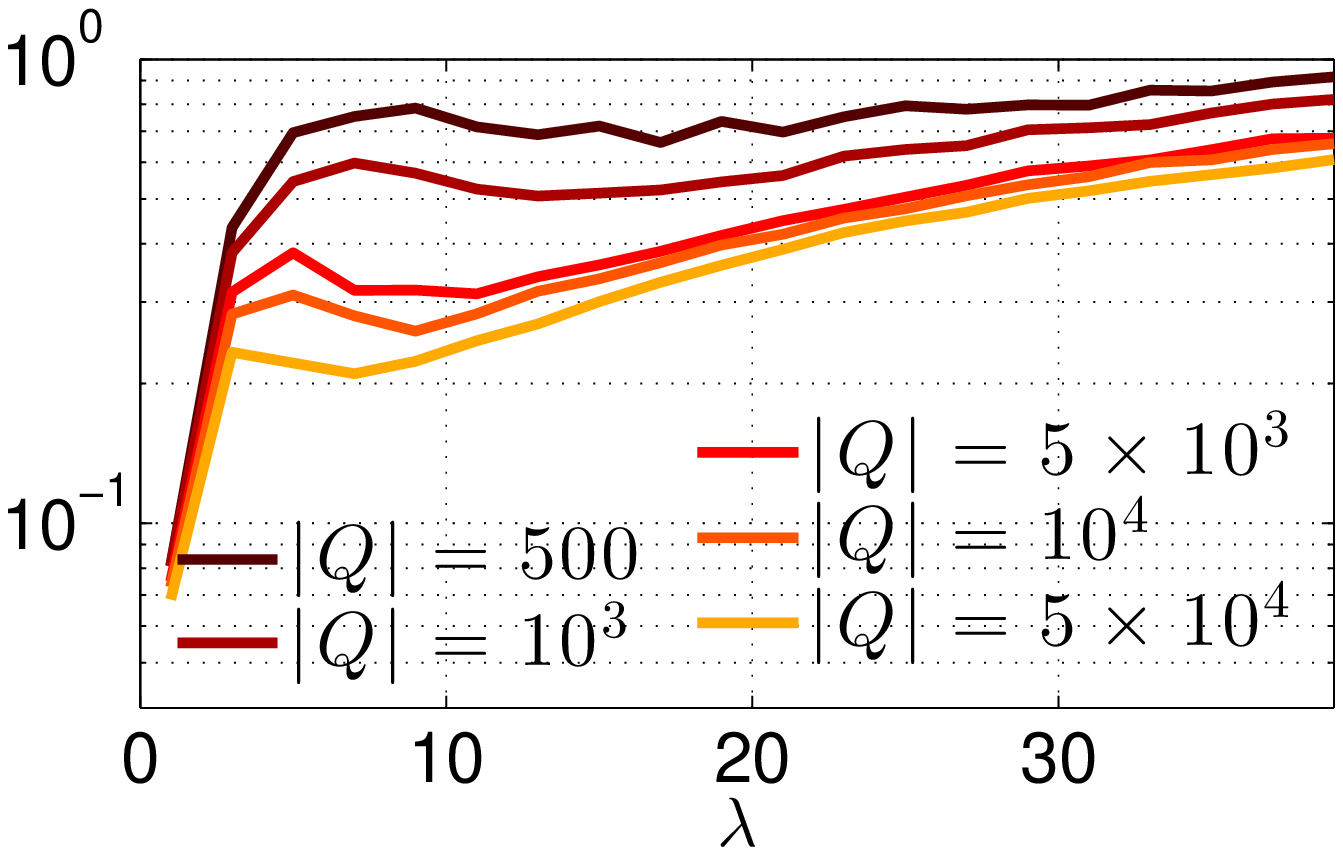}
\includegraphics[scale=.32]{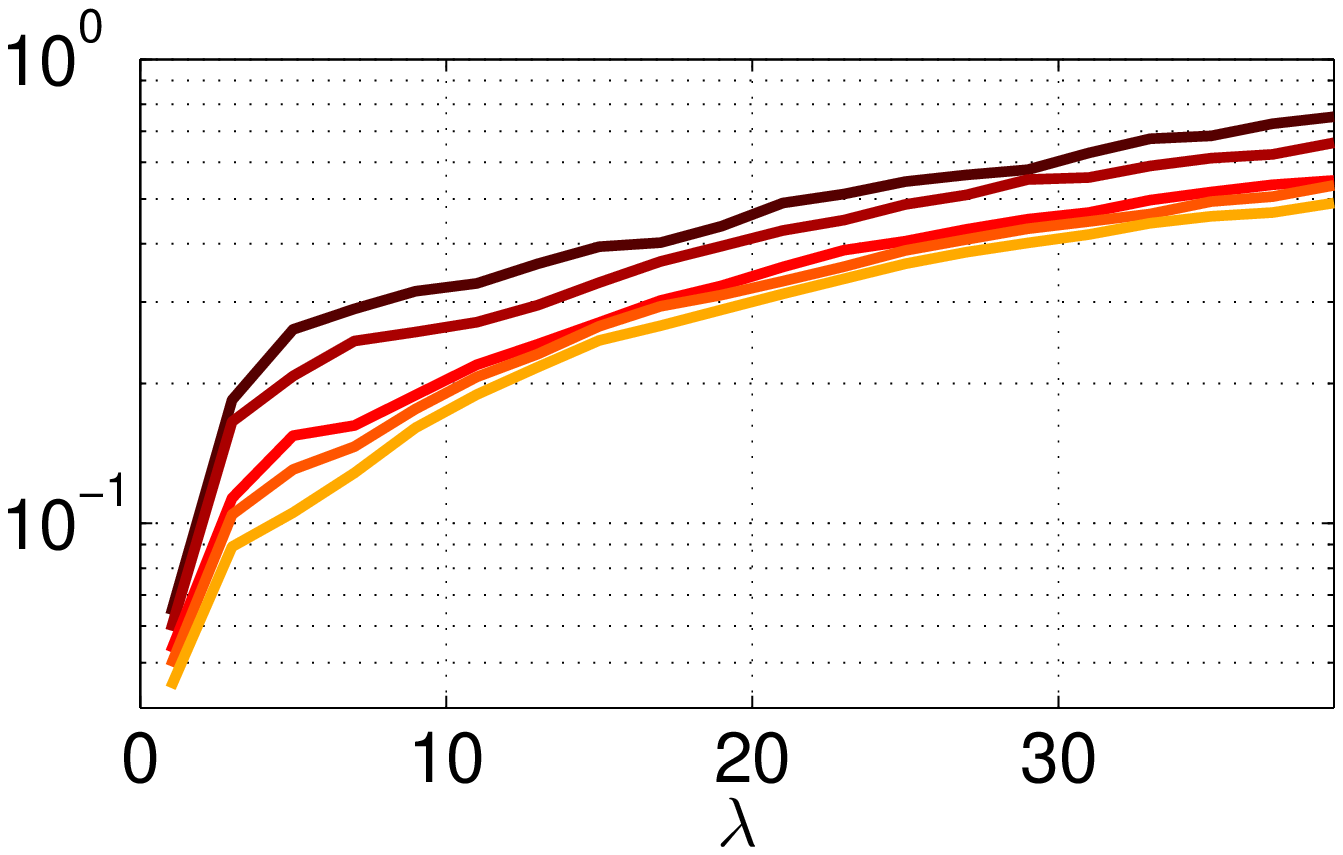}
\caption{\textbf{Estimation performance $\HBx_{\textrm{approx},\mathcal{Q}}$ vs $\HBx$.}  $\mathrm{MSE}(\HBx_{\textrm{approx},\mathcal{Q}},\HBx)$ for different $|\mathcal{Q}|$. SNR is set to $4$dB (resp. $10$dB) on left plot (resp. right plot).\label{fig:approxQuality}}
\vskip .3cm
\centering\includegraphics[scale=.32]{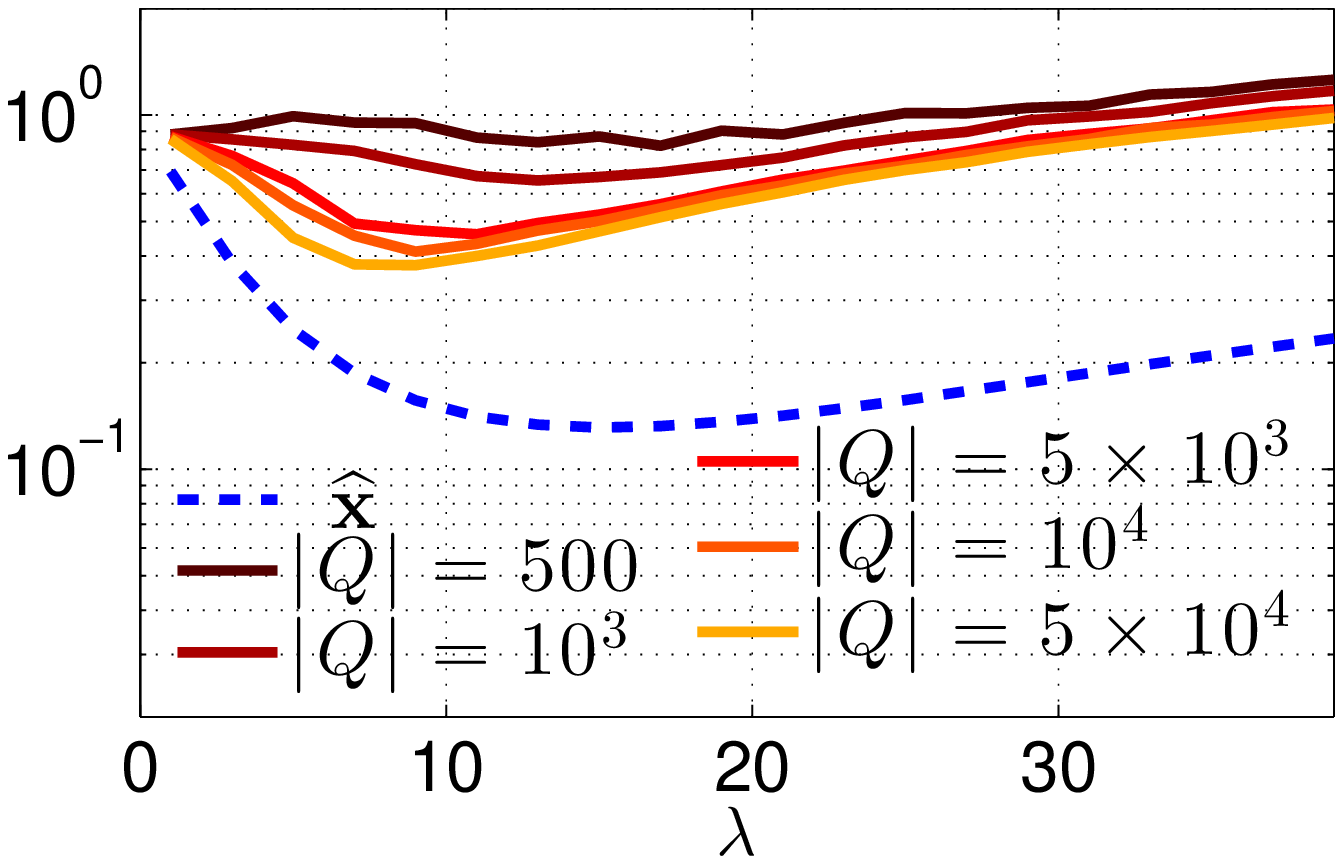}
\includegraphics[scale=.32]{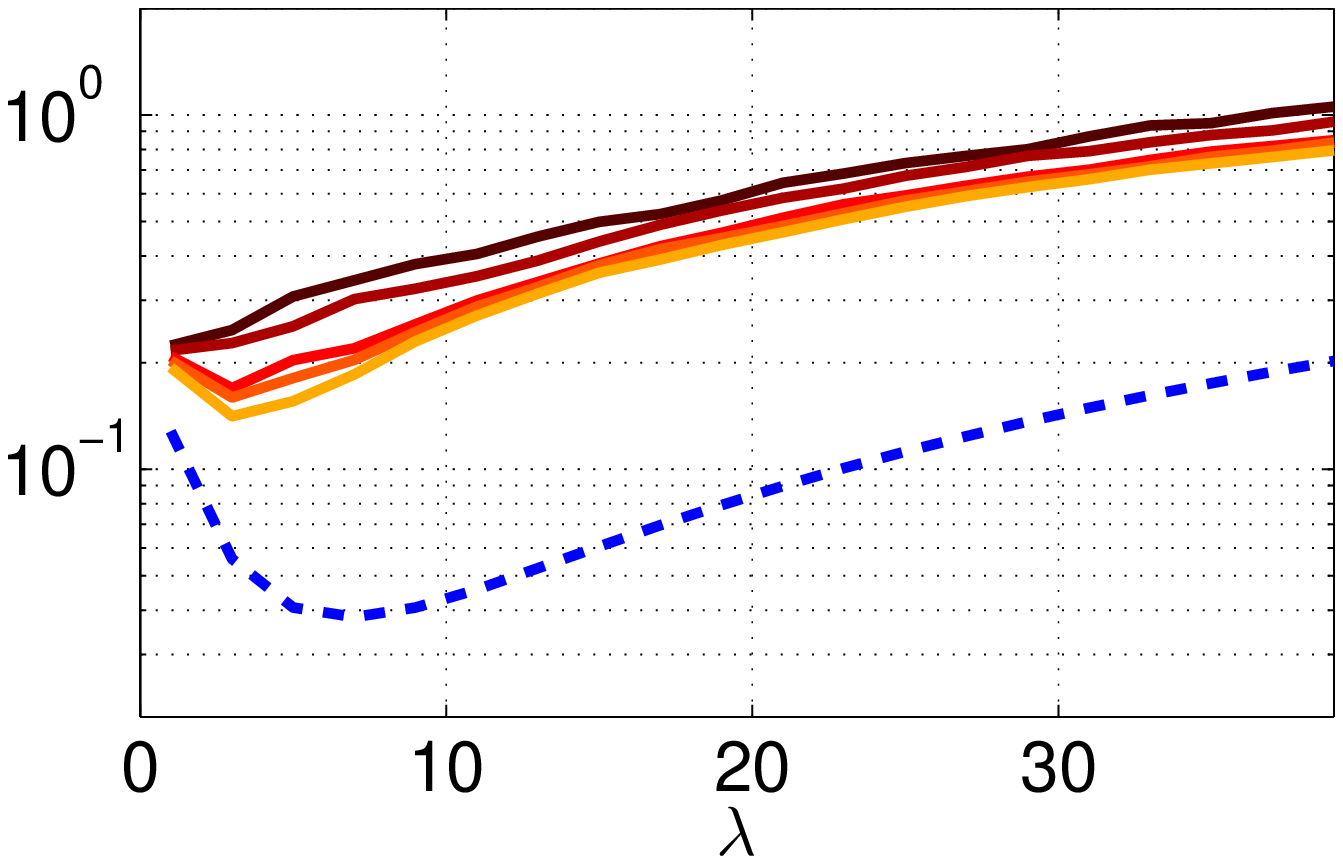}
\caption{\textbf{Estimation performance $\HBx$ vs. $\Bx$ and $\HBx_{\textrm{approx},Q}$ vs. $\Bx$.}  $\mathrm{MSE}(\HBx,\Bx)$ and $\mathrm{MSE}(\HBx_{\textrm{approx},\mathcal{Q}},\Bx)$ for different $|\mathcal{Q}|$.
SNR is set to $4$dB (resp. $10$dB) on left plot (resp. right plot).
\label{fig:estimPerf}}
\end{figure}

\section{Conclusion}

In this contribution, we have developed an algorithm which provides an on-the-fly approximate solution to the multivariate total variation minimization problem. Besides a thorough examination of the KKT conditions, the key-step of the algorithm lies in updating and controlling the range of the upper and lower bounds of the dual solution within a tube of radius $\lambda$. An on-the-fly derivation is achieved by means of an auxiliary vector $\HBz$, which needs to be estimated, providing information on the angle of contact with the tube.
The latter estimation strongly affects the quality of the solution and the proposed on-the-fly estimation of $\HBz$ is currently achieved by assigning a value chosen within a predefined set $\mathcal{Q}$. 
It has been shown that the size of $\mathcal{Q}$ permits to achieve a desired trade-off between the targeted quality of the solution and the application-dependent affordable computational cost. 
In addition, the proposed method could also be extended to other $\ell_{1,p}$ penalization norms in the right-hand side of \eqref{eq:pbPrimal1}, for $p>1$. However one would still face the issue of estimating $\HBz$ which would have to lie within a $\ell_p$ ball of radius $\lambda$. Under current interest is the investigation of how to estimate $\HBz$ in the case where the assumption of piece-wise constant behavior is a priori relaxed.

\section{Appendix}
\begin{figure}[t]
\begin{center}
\includegraphics[scale=.28, clip=true, trim=0cm 0cm 0cm 0cm]{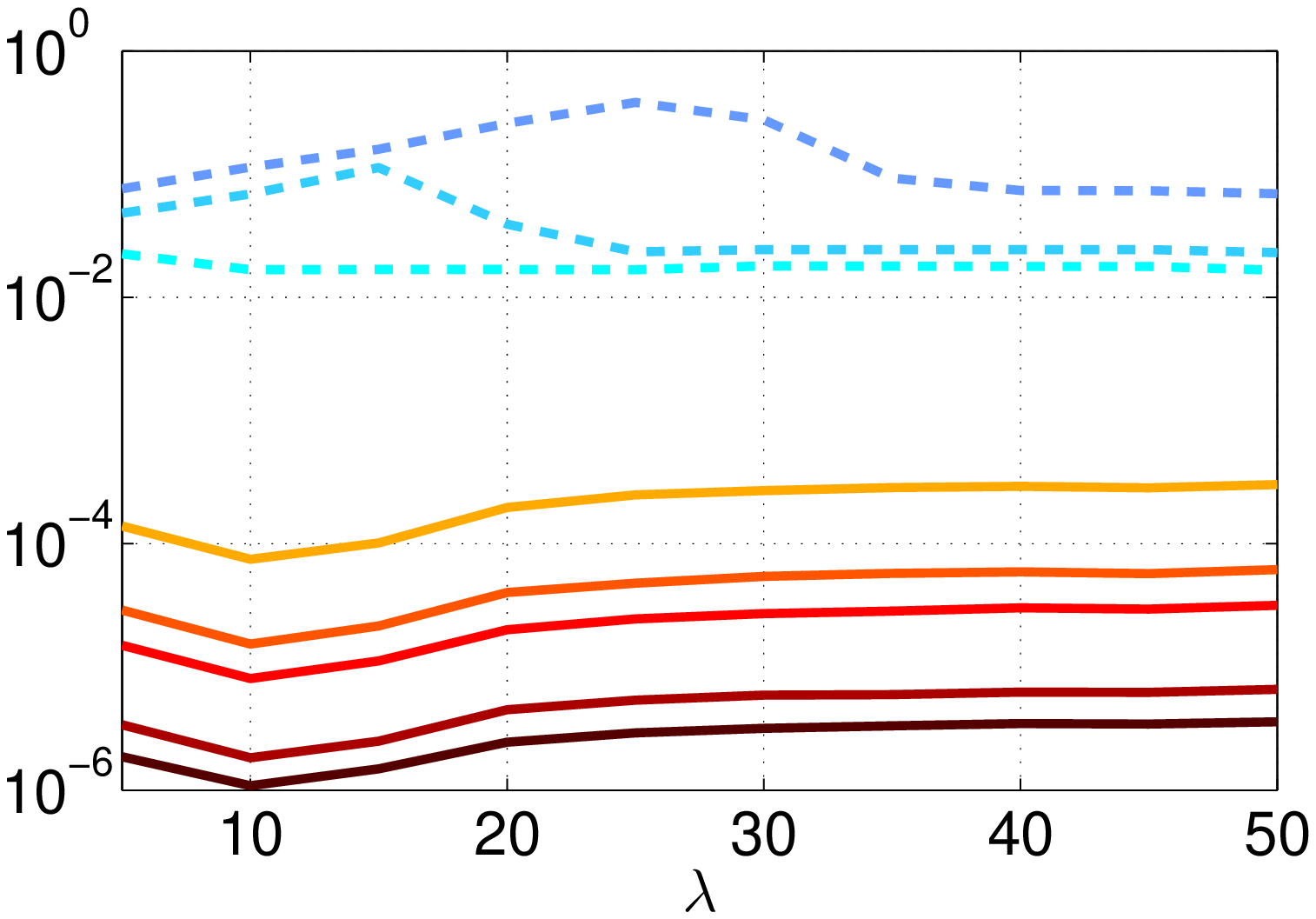}
\includegraphics[scale=.28, clip=true, trim=0cm 0cm 0cm 0cm]{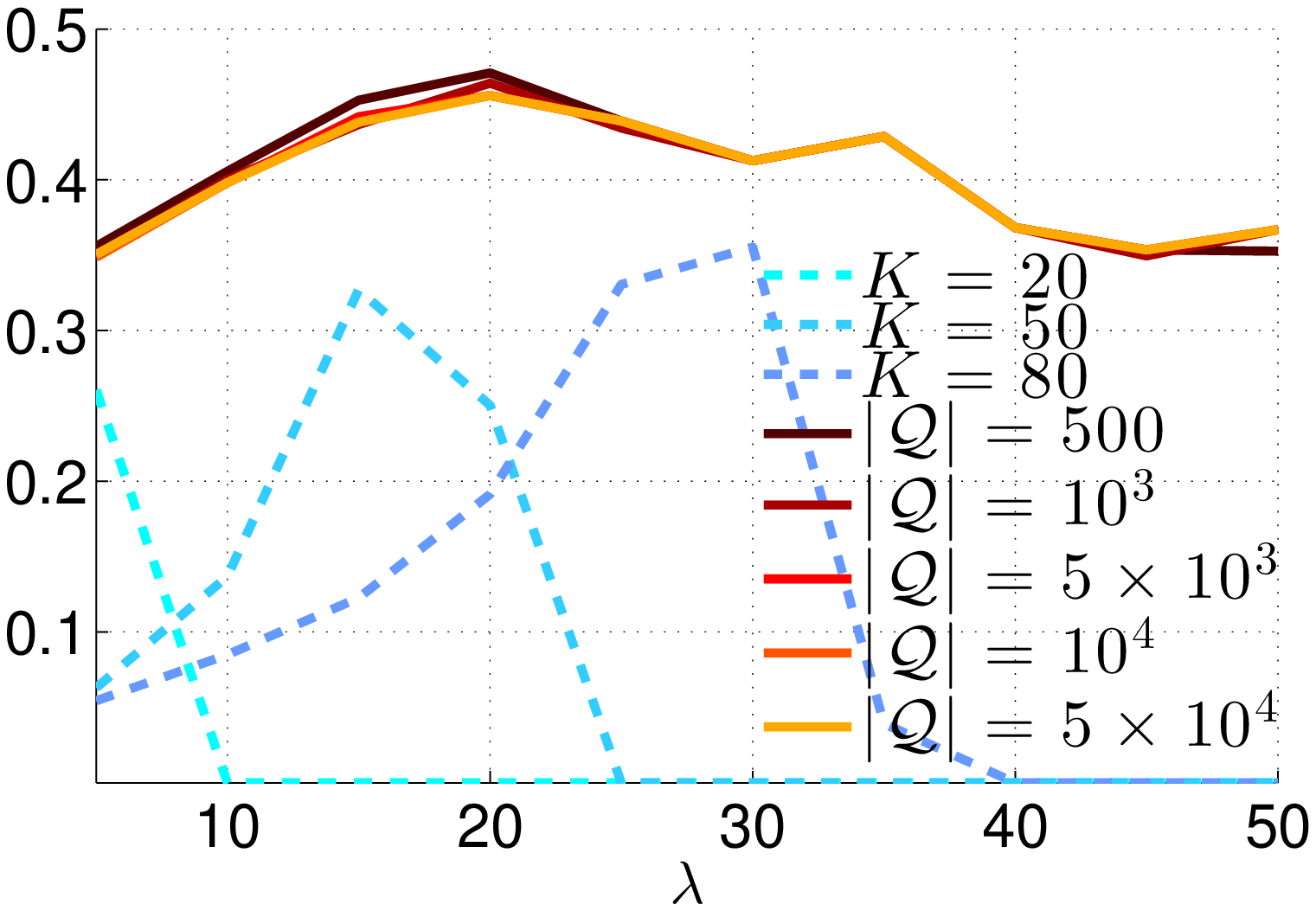}\\
\includegraphics[scale=.28, clip=true, trim=0cm 0cm 0cm 0cm]{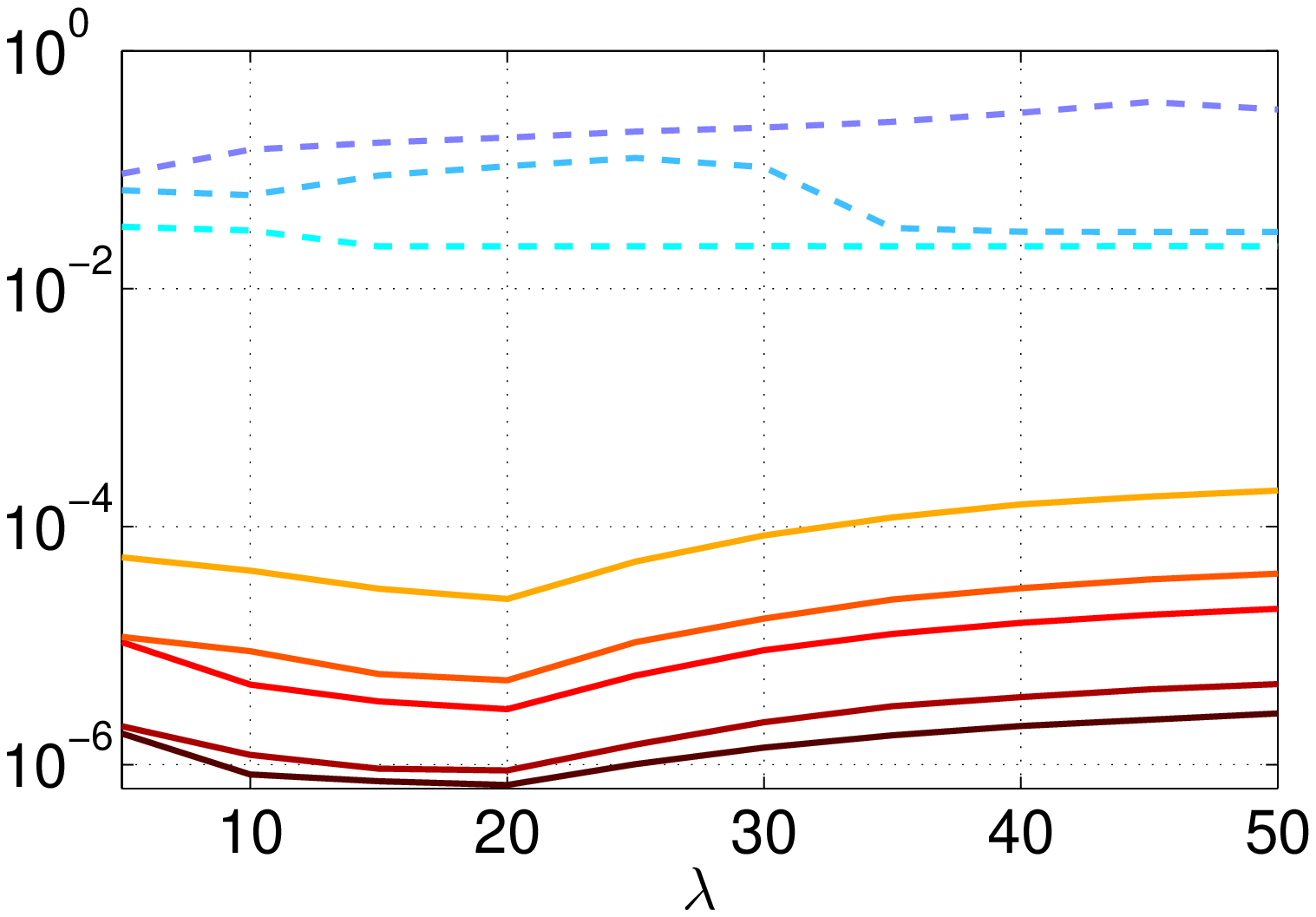}
\includegraphics[scale=.28, clip=true, trim=0cm 0cm 0cm 0cm]{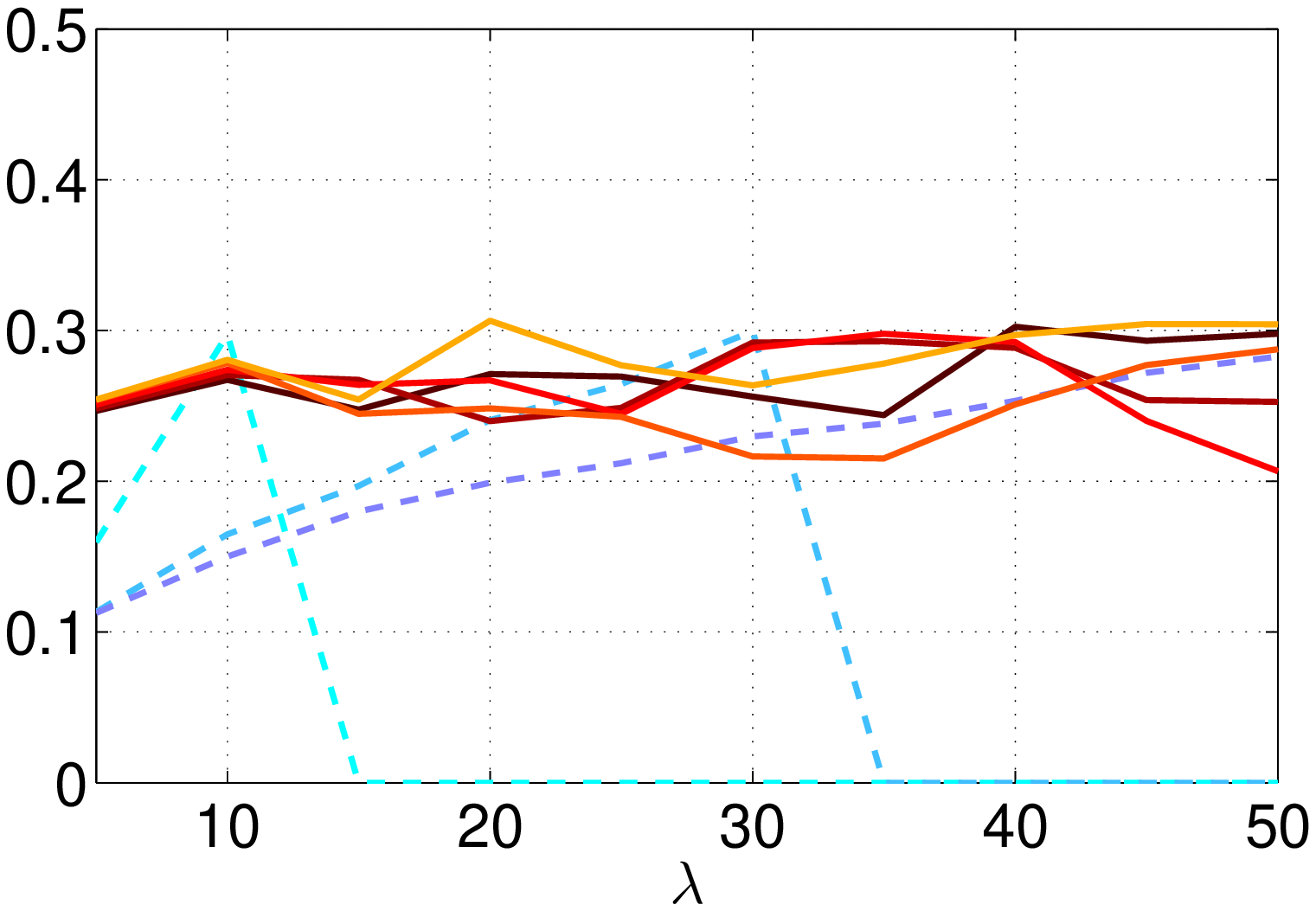}
\end{center}
\caption{\label{fig:onlinePerf}\textbf{Online Performance.} The proposed solution $\HBx_{\mathrm{online},\mathcal{Q}}$ is displayed in solid line while the online ADMM solution $\HBx_{\mathrm{win},K}$ is displayed in dashed line. Performance for $M=2$ (resp. $M=5$) are illustrated top (resp. bottom). Left: median computational cost per incoming sample (in seconds). Right: $J(\widehat{\boldsymbol{r}}_{\mathrm{win},K},\boldsymbol{r})$ and $J(\widehat{\boldsymbol{r}}_{\mathrm{online},\mathcal{Q}},\boldsymbol{r})$ for different values of $|\mathcal{Q}|$ and $K$.}\end{figure}

\subsection{Proof of Equation \eqref{eq:inequaltyDual}}\label{ap:ulb}
According to the primal-dual relation~\eqref{eq:optPrimalDual}, for every $m\in \{1,\ldots,M\}$ and $k\in\{1,\ldots,N-1\}$,
\begin{equation}
\label{eq:pdr}
\widehat{u}_{m,k} = y_{m,k} + u_{m,k-1} - \widehat{x}_{m,k},
\end{equation}
and by definition of the lower and upper bounds of $\widehat{x}_{m,k}$ and $\widehat{u}_{m,k}$, we have 
\begin{align}
\underline{u}_{m,k} &= y_{m,k} + \widehat{u}_{m,k-1} - \underline{x}_{m,k}\label{eq:lb},\\
\overline{u}_{m,k} &= y_{m,k} + \widehat{u}_{m,k-1} - \overline{x}_{m,k}.\label{eq:ub}
\end{align}
By subtracting \eqref{eq:lb} from \eqref{eq:pdr} we obtain
\begin{equation}
\widehat{u}_{m,k} - \underline{u}_{m,k} = \underline{x}_{m,k} - \widehat{x}_{m,k} 
\end{equation}
and, according to \eqref{eq:cond0a}, $\widehat{u}_{m,k} - \underline{u}_{m,k}\leq 0$. The arguments are similar  for proving that $\widehat{u}_{m,k} > \overline{u}_{m,k}$. 
\subsection{Proof of Equation \eqref{eq:updateXmin}\label{app:proofXmin}}

For every $m\in\{1,\ldots,M\}$ and $k\in\{k_0,\ldots,N-2\}$, if 
\begin{equation}
\underline{u}_{m,k+1} = \underline{u}_{m,k}  + y_{m,k+1} - \underline{x}_{m,k} > +\widehat{\constr}_{m,k+1},
\end{equation}
then updating rules of $\underline{x}_{m,k}$, specified in \eqref{eq:majXBi}, have under-evaluated its value $\underline{\nu}_m$. To modify the lower bounds $(\underline{x}_{m,j})_{k_0\leq j \leq k+1}$, on the one hand, we consider the cumulative sum of the observations which, according to \eqref{eq:optPrimalDual}, leads to
\begin{equation}
\sum_{j=k_0+1}^{k+1}y_{m,j} = u_{m,k+1} - u_{m,k_0} + (k - k_0+1) x_{m,k+1},
\end{equation}
and thus, if $\underline{u}_{m,k+1}= + \widehat{\constr}_{m,k+1}$, would lead to
\begin{equation}
\sum_{j=k_0+1}^{k+1}y_{m,j} =\widehat{\constr}_{m,k+1} - u_{m,k_0} + (k - k_0+1)  \underline{\nu}_m,
\label{eq:zcons2}
\end{equation} 
by definition of $\underline{x}_{m,k+1}=\underline{\nu}_m$.
On the other hand, the updating rules \eqref{eq:majUBi} and \eqref{eq:majXBi} have led to
\begin{equation}
\underline{u}_{m,k+1} = \underline{u}_{m,k_0}  + \sum_{j=k_0+1}^{k+1}y_{m,j} - (k - k_0+1)\underline{x}_{m,k}.
\label{eq:zcons3}
\end{equation} 
The combinaison of \eqref{eq:zcons2} and \eqref{eq:zcons3} leads to
\begin{equation}
\underline{\nu}_m = \underline{x}_{m,k} + \frac{- \underline{u}_{m,k_0}  +\underline{u}_{m,k+1} - \widehat{\constr}_{m,k+1} + \widehat{u}_{m,k_0}}{k - k_0+1}.
\end{equation}
Because $\underline{x}_{m,k}$ have been under-evaluated  and by definition $\widehat{u}_{m,k_0} \leq \underline{u}_{m,k_0}$, we can propose the following value
\begin{equation}
\underline{\nu}_m =  \underline{x}_{m,k} + \frac{\underline{u}_{m,k+1} - \widehat{\constr}_{m,k+1} }{k - k_0+1}, 
\end{equation}
in order to adjust the lower bounds, i.e.,
\begin{equation}
(\forall j\in\{k_0,\ldots,k+1\})\quad \underline{x}_{m,j} = \underline{\nu}_m.
\end{equation}
In addition, as a result of $\widehat{u}_{m,k+1}\in[-\widehat{\constr}_{m,k+1},+\widehat{\constr}_{m,k+1}]$ and according to the inequality~\eqref{eq:inequaltyDual}, we set
\begin{equation}
\underline{u}_{m,k+1}=+\widehat{\constr}_{m,k+1}.
\end{equation}

\bibliographystyle{abbrv}
\bibliography{abbr,ref}

\end{document}